\pdfoutput=1
\documentclass[11pt]{article}
\usepackage[]{acl}
\usepackage{times}
\usepackage{latexsym}
\usepackage[T1]{fontenc}
\usepackage[utf8]{inputenc}
\usepackage{microtype}
\usepackage{inconsolata}

\usepackage{kotex}

\usepackage[graphicx]{realboxes}
\usepackage[normalem]{ulem}
\usepackage{cleveref}
\usepackage{adjustbox}
\usepackage{booktabs}
\usepackage{multirow}
\usepackage{amssymb}
\usepackage{hyperref}
\usepackage{subfigure}
\usepackage{arydshln}
\usepackage{xcolor,colortbl}
\usepackage{diagbox}
\usepackage{enumitem}

\crefformat{section}{\S#2#1#3}
\crefformat{subsection}{\S#2#1#3}
\crefformat{subsubsection}{\S#2#1#3}

\title{Translation Deserves Better: Analyzing \\Translation Artifacts in Cross-lingual Visual Question Answering}

\author{ChaeHun Park\thanks{\hspace{0.2cm} Equal contribution}$^1$ \hspace{0.3cm} Koanho Lee\footnotemark[1]$^1$ \hspace{0.3cm} Hyesu Lim$^1$ \hspace{0.3cm} Jaeseok Kim$^2$ \\
\hspace{0.3cm}    \textbf{Junmo Park}$^2$\hspace{0.3cm}    \textbf{Yu-Jung Heo}$^2$\hspace{0.3cm}    \textbf{Du-Seong Chang}$^2$ \hspace{0.3cm} \textbf{Jaegul Choo}$^1$\\
$^1$ KAIST AI \hspace{0.3cm} $^2$ KT Corporation\\
\hspace{0cm}\texttt{\{ddehun,le5544,hyesulim,jchoo\}@kaist.ac.kr} \\
\hspace{0cm}\texttt{\{jaeseok.kim,jm.p,yj.heo,dschang\}@kt.com} \\
 }

\begin{document}
\maketitle
\begin{abstract}
Building a reliable visual question answering~(VQA) system across different languages is a challenging problem, primarily due to the lack of abundant samples for training.
To address this challenge, recent studies have employed machine translation systems for the cross-lingual VQA task. 
This involves translating the evaluation samples into a source language (usually English) and using monolingual models (i.e., \textit{translate-test}). 
However, our analysis reveals that translated texts contain unique characteristics distinct from human-written ones, referred to as \textit{translation artifacts}.
We find that these artifacts can significantly affect the models, confirmed by extensive experiments across diverse models, languages, and translation processes.
In light of this, we present a simple data augmentation strategy that can alleviate the adverse impacts of translation artifacts.

\end{abstract}
\section{Introduction}
Visual question answering~(VQA) aims to answer an open-ended question by reasoning about a given image~\citep{Agrawal2015VQAVQ}. 
Despite recent advances in vision-language~(VL) modeling, building proficient models across various languages is still challenging. 
This issue primarily arises from the limited availability of annotated datasets, which are predominantly in high-resource languages such as English.
Although recent efforts in developing multilingual VL models can address this issue to some extent \citep{uc2,xuniter,li-etal-2023-unifying, geigle2023mblip}, training on datasets in the target languages is still crucial for enhanced model performance in those languages.

To mitigate the data scarcity issue, cross-lingual transfer learning focuses on utilizing extensive datasets in a \textit{source} language~(typically English) to build models effective in a \textit{target} language~\cite{artetxe-etal-2020-translation, bugliarello-etal-2022-iglue}.
One of the popular approaches, namely \textit{translate-train}, translates training samples into individual target languages and uses them to train models for target languages. 
This approach is advantageous as it does not perform translation during inference, but it requires training individual models for each target language. 
Furthermore, recent VL models~\citep{flava,llava,BLIP2}, which are mostly tailored in English, are not suitable for the translate-train approach.
Another widely adopted approach, called \textit{translate-test}, translates test samples written in target languages into the source language and uses VL models of the source language for the inference. 
These translation-based approaches have shown remarkable performance in cross-lingual tasks.

\begin{figure}[t!]
\centering
\includegraphics[width=0.95\columnwidth]{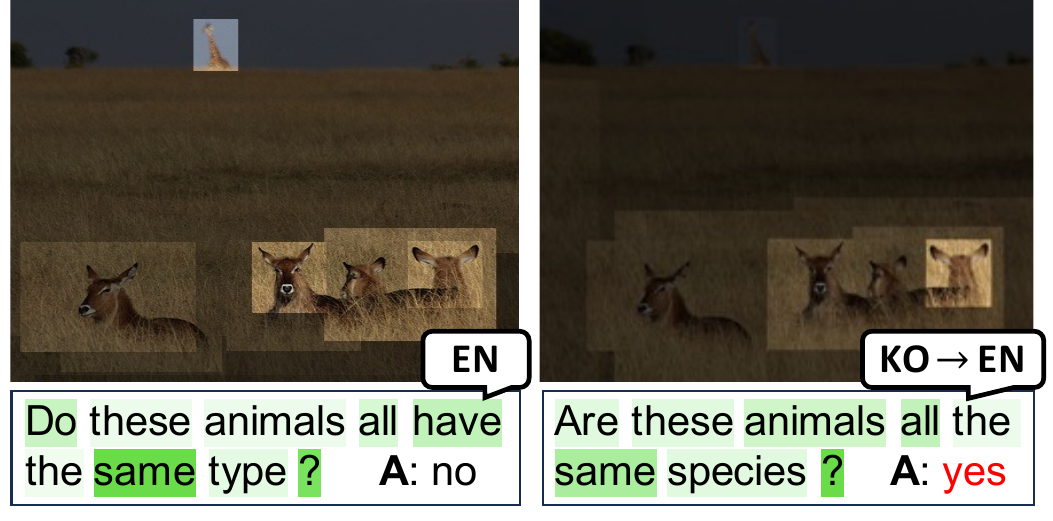}
\caption{Predictions of LXMERT~\citep{LXMERT} on the \textbf{original}~(left) and \textbf{translated}~(right) questions. 
The model is correct for the human-written question but is incorrect for the correctly translated one.
The original Korean question is \textit{``이 동물들은 모두 같은 종입니까?''}.
For model visualization, we use an attention-based method by \citet{Attentionvisualization}.}
\vspace{-4mm}
\label{fig:motif}
\end{figure}

Despite the effectiveness of translation systems in cross-lingual VL tasks, using machine-translated texts as input inevitably introduces a mismatch between the training and inference phases.
In the translate-test approach, models are trained on human-written texts but evaluated on machine-translated texts. 
This distribution shift could hurt the generalization of models to different languages~\citep{yu-etal-2022-translate,CLS_translationese}.
For instance, as illustrated in Fig.~\ref{fig:motif}, leveraging machine-translated texts might lead to undesirable model outcomes, even when both questions convey the same meaning.
In this paper, we refer to artifacts in translations that cause such unwanted behaviors as \textit{translation artifacts}.
We argue that the translation artifacts have been overlooked in previous cross-lingual VQA studies despite their significance.

To explore the effect of mismatched data distribution on cross-lingual VQA, we alleviate this mismatch in the data origins\footnote{We refer to \textit{origin} as a writer of texts~(\textit{i.e.}, human or machine translation system).} by employing machine-translated texts in both training and inference.
Our investigation focuses on the translate-test, which can take advantage of strong monolingual models and efficiently serve multiple target languages with a single VL model.
Our results reveal that models trained on machine-translated texts generally outperform those trained on human-written texts, increasing the averaged accuracy over languages and models from 51.82 to 53.14 points.
This improvement, as confirmed by our qualitative analysis, is primarily attributed to the subtle nuances in translated texts~(\textit{i.e.}, translation artifacts).
Our comprehensive study covers various components in cross-lingual VQA, including 14 models, 13 languages, 5 machine translation systems, and diverse translation setups. 
We also observe that recent VL models~\citep{BLIP2, instructBLIP} integrated with large language models also suffer from translation artifacts. 
Finally, we present simple data augmentation techniques, verifying their effectiveness in both human-written and translated texts.

Our contribution can be summarized as follows:
\begin{enumerate}
\itemsep0.3em 
\item This is, to our knowledge, the first study to investigate translation artifacts in cross-lingual visual question answering.
\item We provide extensive analyses across a variety of languages and models, providing a foundation for future research.
\item We present simple yet effective data augmentation strategies using translated texts.
\end{enumerate}

\section{Related Work}
\label{sec:related_work}

\subsection{Cross-lingual VQA}
The study of VQA has predominantly focused on English and other high-resource languages~\citep{Zhu2015Visual7WGQ, Agrawal2015VQAVQ, Goyal2016MakingTV, Marino2019OKVQAAV, a-okvqa}.
To extend the use of VQA to various languages, researchers have introduced cross-lingual transfer techniques~\citep{m3p, uc2, Nooralahzadeh2022ImprovingTC, liu2023delving}.
One effective approach involves pretraining VL models on multilingual image-text pairs and then fine-tuning them on English VQA, which is known as \textit{zero-shot} transfer~\citep{jain-etal-2021-mural-multimodal, lee-etal-2022-efficient, Zeng2022CrossViewLM, pali, chen-etal-2023-mclip, li-etal-2023-unifying}.
Another popular approach that leverages advanced machine translation shows promise in adapting to various languages. 
The \textit{translate-train} involves translating the text pairs from high-resource languages to the target language for finetuning~\citep{thapliyal-soricut-2020-cross, Zeng2022CrossViewLM, chen-etal-2023-mclip, li-etal-2023-unifying}. 
On the other hand, the \textit{translate-test} uses machine translation to convert test data into English, allowing the use of English-only models for inference~\citep{jain-etal-2021-mural-multimodal, bugliarello-etal-2022-iglue, xgqa}. 
This latter approach is particularly beneficial, considering the strong performance of existing English-only models~\citep{flava, BLIP2, instructBLIP, llama_adapter_v2}.

\subsection{Translation Artifacts}
Translated texts often exhibit unique characteristics, referred to as \textit{translation artifacts} or \textit{translationese}~\citep{Gellerstam1986TranslationeseIS, lembersky2012adapting, baker2019corpus, edunov-etal-2020-evaluation}. 
These characteristics can negatively influence model outcomes due to their stylistic deviations from the original texts~\citep{volansky2015features, bizzoni-etal-2020-human, yu-etal-2022-translate}.
~\citet{yang2021enhancing} examined the representation discrepancies between English and other languages in the translate-train approach for various language understanding tasks. 
~\citet{CLS_translationese} explored the effects of translation artifacts on model evaluation in cross-lingual summarization.
To mitigate the effects of translation artifacts, researchers have proposed various methods, such as incorporating machine-translated sentences in training~\citep{artetxe-etal-2020-translation, yu-etal-2022-translate, CLS_translationese} or utilizing specific tags to differentiate between original and machine-translated texts~\citep{marie2020tagged, riley-etal-2020-translationese, wang-etal-2021-language-coverage}.

However, the effect of translation artifacts on cross-lingual VQA remains largely underexplored, leading to potential risks and unexpected outcomes. 
While previous research has primarily focused on the application of machine translation in VL models~\citep{thapliyal-soricut-2020-cross, Zeng2022CrossViewLM, bugliarello-etal-2022-iglue, xgqa, maxm, chen-etal-2023-mclip}, our study aims to identify the presence and impact of translation artifacts within cross-lingual VQA. 
We find that these translation artifacts are prevalent in VL models handling both image and text modalities. 
\section{Translation Artifacts in Cross-lingual Visual Question Answering}
\label{sec:analysis}

\begin{figure*}[t!]
\centering
\includegraphics[width=0.98\textwidth]{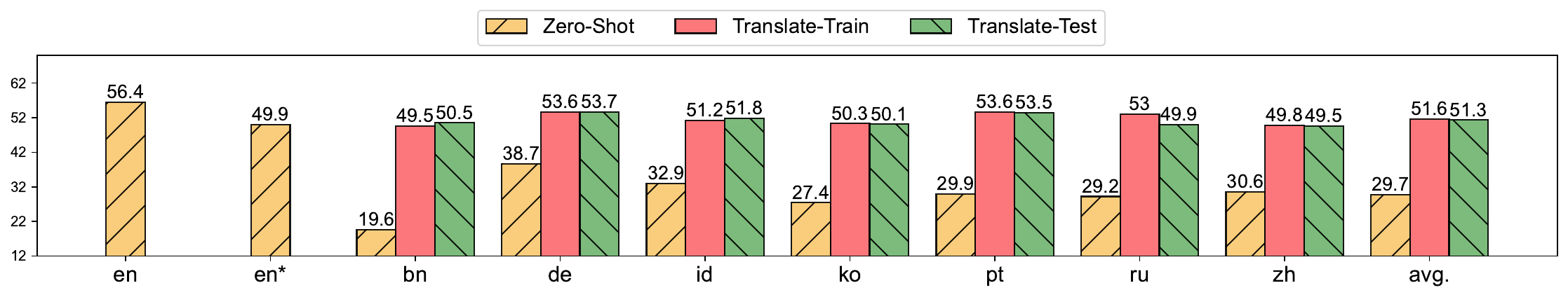}
\vspace{-3mm}
\caption{\textbf{Averaged multilingual models results} 
The \textit{en*} and \textit{avg.} denote the RT-translated English evaluation set and the averaged cross-lingual transfer results, respectively.  
Full results of each multilingual model are in Fig.~\ref{fig:multilingual_full}.}
\label{fig:multilingual_avg}
\end{figure*}
\begin{table*}[t!]
    \tiny
    \centering
    
    \begin{adjustbox}{width=0.98\textwidth}
        \begin{tabular}{ccc|ccccccccc}
            \midrule
            \textbf{Models} & \multicolumn{1}{l}{\textbf{RT?}} & \multicolumn{1}{c}{en} & en$^*$ & bn & de & id & ko & pt & ru & zh & \textbf{avg.}\\ \noalign{\vskip 0.15ex}
            \hline         
             &  & \underline{57.33} & 50.14 & 50.67 & 54.09 & 52.54 & 50.67 & 54.21 & 49.69 & 49.57 & 51.63 \\
            \rowcolor[rgb]{0.882,0.882,0.882} \multirow{-2}{*}{MUNITER} \cellcolor[rgb]{1,1,1} &  \checkmark  & 55.70 & \underline{ 52.75} & \underline{52.34} & \underline{55.66} & \underline{53.48} & \underline{53.36} & \underline{54.72} & \underline{53.98} & \underline{52.29} & \underline{53.69} \\ 
             
             &  & \underline{56.98} & 49.90 & 50.76 & 54.63 & 52.37 & 50.52 & \underline{54.24} & 48.91 & 49.94 & 51.62 \\
             \rowcolor[rgb]{0.882,0.882,0.882} \multirow{-2}{*}{XUNITER} \cellcolor[rgb]{1,1,1}&  \checkmark  &55.22 & \underline{ 52.45}  & \underline{52.10} & \underline{54.97} & \underline{52.66} & \underline{52.51} & 54.18 & \underline{52.85} & \underline{52.23} & \underline{53.07}\\
             
            &  & \underline{56.85} & 50.22 & 51.34 & 54.01 & 52.35 & 50.75 & 53.81 & 51.93 & 50.04 & 52.03 \\
             \rowcolor[rgb]{0.882,0.882,0.882} \multirow{-2}{*}{UC$^2$} \cellcolor[rgb]{1,1,1}&  \checkmark  & 55.12 & \underline{52.44} & \underline{52.35} & \underline{55.10} & \underline{53.29} & \underline{53.07} & \underline{54.17} & \underline{53.36} & \underline{52.73} & \underline{53.44}\\
             
             &  & \underline{54.45} & 49.29 & 49.18 & 52.14 & 49.87 & 48.59 & 51.87 & 49.05 & 48.38 & 49.87 \\
              \rowcolor[rgb]{0.882,0.882,0.882} \multirow{-2}{*}{M$^3$P}\cellcolor[rgb]{1,1,1}&  \checkmark  & 52.97 & \underline{51.97} & \underline{50.63} & \underline{53.03} & \underline{51.42} & \underline{50.38} & \underline{52.11} & \underline{51.80} & \underline{50.41} & \underline{51.40}\\
             
             &  & \underline{55.40} & 48.42 & 49.64 & 52.83 & 50.80 & 49.17 & \underline{52.49} & 47.54 & 48.02 & 50.07 \\
             \rowcolor[rgb]{0.882,0.882,0.882} \multirow{-2}{*}{LXMERT} \cellcolor[rgb]{1,1,1}&  \checkmark  & 53.44 & \underline{50.51} & \underline{50.20} & \underline{52.93} & \underline{51.34} & \underline{50.41} & 52.47 & \underline{51.44} & \underline{50.25} & \underline{51.29}\\
             
             &  & \underline{57.47} & 50.11 & 51.74 & 54.52 & 52.79 & 51.27 & 54.56 & 52.27 & 50.33 & 52.50 \\
             \rowcolor[rgb]{0.882,0.882,0.882}\multirow{-2}{*}{UNITER}\cellcolor[rgb]{1,1,1}&  \checkmark  & 55.92 & \underline{52.90} & \underline{52.32} & \underline{55.53} & \underline{53.67} & \underline{52.93} & \underline{54.66} & \underline{53.56} & \underline{52.60} & \underline{53.61}\\
             
            &   & \underline{56.72} & 50.10 & 50.84 & 54.10 & 52.27 & 50.73 & 53.98 & 49.91 & 49.92 & 51.68 \\
             \rowcolor[rgb]{0.882,0.882,0.882}\multirow{-2}{*}{VILBERT} \cellcolor[rgb]{1,1,1}&  \checkmark  & 55.22 & \underline{52.52} & \underline{52.23} & \underline{54.85} & \underline{53.43} & \underline{52.75} & \underline{54.26} & \underline{53.69} & \underline{52.22} & \underline{53.35}\\
             
            &   & \underline{55.17} & 48.66 & 49.43 & 52.58 & 50.34 & 48.66 & \underline{52.72} & 50.50 & 48.89 & 50.45 \\
             \rowcolor[rgb]{0.882,0.882,0.882}\multirow{-2}{*}{VisualBERT} \cellcolor[rgb]{1,1,1}&  \checkmark  & 53.51 & \underline{50.91} & \underline{50.57} & \underline{53.10} & \underline{51.17} & \underline{50.45} & 52.59 & \underline{51.47} & \underline{50.97} & \underline{51.47}\\
             
             &   & \underline{57.79} & 50.32 & 51.22 & 54.47 & 52.62 & 50.94 & \underline{54.79} & 51.17 & 50.02 & 52.18 \\
             \rowcolor[rgb]{0.882,0.882,0.882}\multirow{-2}{*}{VL-BERT}\cellcolor[rgb]{1,1,1}&  \checkmark  & 55.61 & \underline{52.79} & \underline{52.38} & \underline{55.27} & \underline{53.43} & \underline{52.58} & 54.63 & \underline{53.32} & \underline{52.31} & \underline{53.42}\\
             
            &  & \underline{58.05} & 52.10 & 52.03 & 54.70 & 52.99 & 51.57 & 54.91 & 52.36 & 51.22 & 52.83 \\
             \rowcolor[rgb]{0.882,0.882,0.882}\multirow{-2}{*}{BLIP-2} \cellcolor[rgb]{1,1,1}&  \checkmark  & 56.11 & \underline{54.76} & \underline{53.18} & \underline{55.70} & \underline{53.98} & \underline{53.51} & \underline{55.11} & \underline{54.25} & \underline{53.31} & \underline{54.15}\\
             
             &  & \underline{57.85} & 52.26 & 51.80 & 54.91 & 53.01 & 51.29 & \underline{54.85} & 53.16 & 51.34 & 52.91 \\
             \rowcolor[rgb]{0.882,0.882,0.882}\multirow{-2}{*}{InstructBLIP}\cellcolor[rgb]{1,1,1}&  \checkmark  & 55.84 & \underline{54.62} & \underline{53.04} & \underline{55.06} & \underline{53.82} & \underline{53.17} & 54.32 & \underline{54.08} & \underline{53.18} & \underline{53.81}\\
             
             &  & \underline{\textbf{58.84}} & 52.91 & 53.47 & 56.26 & 54.11 & 52.85 & 55.84 & 53.64 & 52.18 & 54.05 \\
             \rowcolor[rgb]{0.882,0.882,0.882}\multirow{-2}{*}{FLAVA}\cellcolor[rgb]{1,1,1}&  \checkmark  & 56.87 & \underline{\textbf{55.07}} & \underline{\textbf{53.94}} & \underline{\textbf{56.35}} & \underline{\textbf{54.99}} & \underline{\textbf{54.51}} & \underline{\textbf{55.96}} & \underline{\textbf{55.61}} & \underline{\textbf{53.82}} & \underline{\textbf{55.03}}\\ 
            
             &  & \underline{ 56.91} & 50.37 & 51.01 & 54.10 & 52.17 & 50.58 & 54.02 & 50.84 & 49.99 & 51.82 \\
            \rowcolor[rgb]{0.882,0.882,0.882}\multirow{-2}{*}{\textbf{avg.}} \cellcolor[rgb]{1,1,1}&  \checkmark & 55.13 & \underline{52.81} & \underline{52.11} & \underline{54.80} &  \underline{53.06} & \underline{52.47} &  \underline{54.10} &  \underline{53.28} & \underline{52.19} &  \underline{53.14}\\
             \toprule
            \end{tabular}
        
    \end{adjustbox}
    \caption{\textbf{Translate-test results with different origins of training dataset}
    For languages other than English, we use an evaluation set released by \citet{bugliarello-etal-2022-iglue} translated with Google Machine Translation~(GMT).         
    Here, \textit{en$^*$} denotes the RT-translated English evaluation set. 
    Models finetuned on RT-translated English texts are marked with \checkmark. 
    For each model within the different data origins, the higher score in each column is highlighted in \underline{underline}.
    The highest score in each column is further highlighted in \textbf{bold}.
    The statistical significance analysis is in Appendix~\ref{sec: Appendix Statistical Test}.
    }
    \label{tab:translate-test_main}
    \vspace{-3mm}
    \end{table*}

In this work, we analyze the impact of machine translation on cross-lingual VQA tasks, especially on the translate-test approach. 
To this end, we vary the origin of training datasets into human and a machine translation~(MT) system and then observe how this change affects the model behavior. 
We use roundtrip~(RT) translation to generate machine-translated training samples from the source language- English.\footnote{Afterward, the \textit{source} language refers to English.}

\subsection{Experimental Setup}
\label{secsec:setup}

\subsubsection{Data}
We use xGQA~\citep{xgqa}, a representative benchmark for the cross-lingual VQA task. 
Each sample in the dataset consists of an image, a structured question related to the image, and an answer. 
The training set is derived from the original English GQA dataset~\citep{hudson2019gqa} and consists of 72k images and 943k samples. 
The evaluation sets cover seven different languages - Bengali~(bn), German~(de), Indonesian~(id), Korean~(ko), Mandarian~(zh), Portuguese~(pt), and Russian~(ru) - and is manually translated from the balanced test-dev set of the English GQA dataset by human annotators.
The evaluation set consists of 398 images and 12,578 samples, and all images in xGQA datasets are sampled from the Visual Genome dataset~\citep{visualgenome}. 
Further details on the dataset are described in Appendix~\ref{sec: Appendix xGQA details}.

\subsubsection{Models}
\label{secsec:model}
We conduct experiments with all multilingual and monolingual VL models addressed in \citet{bugliarello-etal-2022-iglue}. 
Specifically, for multilingual models, MUNITER~\citep{xuniter}, XUNITER~\citep{xuniter}, UC$^2$~\citep{uc2}, and M$^3$P~\citep{m3p} are used.
For monolingual English-only models, LXMERT~\citep{LXMERT}, UNITER~\citep{chen2019uniter}, VILBERT~\citep{lu2019vilbert}, VisualBERT~\citep{li2019visualbert}, and VL-BERT~\citep{VL_BERT} are used. 
All models are based on transformer~\citep{vaswani2017attention} architecture, and both image and text are fed to the network simultaneously. 
In addition, we conduct experiments with recently proposed monolingual English VL models - BLIP-2~\citep{BLIP2}, InstructBLIP~\citep{instructBLIP}, and FLAVA~\citep{flava}.
More details are in Appendix~\ref{sec: Appendix Model details}.

For the cross-lingual transfer of multilingual models, the following approaches are considered: zero-shot, translate-train, and translate-test. 
The zero-shot approach trains a model on the original English training set in the GQA dataset and directly uses it to infer evaluation samples in the target language.\footnote{The \textit{zero-shot} denotes that the language of evaluation samples differs from the language used in the finetuning phase.} 
The translate-train approach trains individual models for each target language on a translated training dataset.
The translate-test approach trains a single model on an English training dataset and uses it for the evaluation of target languages along with a translation system.
For monolingual models, only the translate-test approach is evaluated.

\subsubsection{Training Dataset from Different Origins}
For the translate-test, we finetune all models described above on English GQA datasets from two different origins individually: Human and MT. 
For Human, we use the original xGQA training set. 
For MT, we use the roundtrip (RT) translation to obtain training samples that are written by an MT system. 
We use \textsc{nllb}~\citep{NLLB} as the MT system for RT translation.\footnote{\url{facebook/nllb-200-3.3B} is used.} 
The German~(de) is used as a pivot language during RT translation~(en$\rightarrow$de$\rightarrow$en).
For the zero-shot and translate-train, we use the original English dataset and the dataset translated from English to individual target languages, respectively.
More details about translation processes are in Appendix~\ref{sec: Appendix Translation}.

\subsubsection{Evaluation dataset}
\noindent \textbf{Source Language} 
For English evaluation, we use the official evaluation set released by \citet{xgqa}~(en).
Besides, we also make translated versions of English evaluation sets through RT translation~(en$^*$). 
This process is to understand the impact of data origins on models more comprehensively. 

\noindent \textbf{Target Languages}
For zero-shot and translate-train evaluations, the target language questions released by \citet{xgqa} are used.
For the translate-test evaluation, each question in the target language should be translated into English. 
In this work, we use an official translate-test evaluation set~\citep{bugliarello-etal-2022-iglue} generated by the Google Machine Translation (GMT) system.

\subsubsection{Implementation Details}
\label{secsecsec: implementation details}
For finetuning VL models, we follow hyperparameters reported in \citet{bugliarello-etal-2022-iglue} for a fair comparison. 
Specifically, all models are trained for 5 epochs, and the batch size and initial learning rate are set to 256 and 4e-5, respectively. 
AdamW~\citep{adamw} is used for optimization.
All models are trained with a classification head on top of image-language representation.
We evaluate models after every training epoch and choose the best checkpoint based on its accuracy on the original English development set. 
More implementation details are in Appendix~\ref{sec: Appendix Implementation details}.

\section{Results and Analysis}
\label{sec:results_and_analysis}

\subsection{Main Results}
\label{secsec:results}
\noindent \textbf{Multilingual Models} Fig.~\ref{fig:multilingual_avg} presents averaged evaluation results of multilingual models with different cross-lingual transfer approaches.
The models show decreased accuracy when transferred to languages other than English.
For instance, the average accuracy is 56.4 for the original English dataset, but are 51.6 and 51.3 for translate-train and translate-test approaches, respectively. 
Among the different cross-lingual transfer approaches, translate-train and translate-test are comparable, while the zero-shot approach usually performs worse.

\noindent \textbf{Misaligned Data Origins in Translate-Test} 
Table~\ref{tab:translate-test_main} presents translate-test evaluation results of models with different training data origins. 
Regarding models trained on human texts, FLAVA usually performs better than other models. 
Regarding the effects of different training data origins, we observe that models generally show higher accuracy when the origins of training and evaluation datasets are matched. 
Specifically, for the original English evaluation set, models trained on human texts consistently perform better than ones trained on MT texts. 
On the contrary, for the translate-test, in which all questions are generated by MT systems, models trained on MT texts outperform those trained on human texts. 
By only aligning the data origins of training and evaluation sets, the averaged translate-test scores across models and languages are increased from 51.82 to 53.14. 
Note that this trend is consistent in RT-translated English evaluation set~(en$^*$), where the average score increases from 50.37 to 52.81. 
Based on our results, we suggest a reconsideration of factors contributing to lower scores of target languages in cross-lingual VQA, indicating that data origin misalignment, alongside translation errors, could significantly impact the success of language transfer.

\subsection{Human Analysis}
\label{secsec:human_analysis}

We next analyze translated questions in the evaluation set to examine where the increased performance of models trained on MT texts comes from. 
To this end, we annotate translate-test evaluation samples in which a model trained on human texts makes wrong predictions, but a model trained on MT texts makes correct ones. 
Note that we only consider the translate-test samples in which both models with human and MT texts correctly predicted the paired human-written English samples to avoid wrong predictions arising from sample complexity.
UC$^2$ is selected as a VQA model, and 200 questions from the Korean (ko) evaluation set are annotated. 
Two native speakers annotate the MT errors in each question, and the annotation schema is based on multidimensional quality metric~(MQM) ontology~\citep{MQM} following~\citet{moghe-etal-2023-extrinsic}. 
More details about human annotation and annotated examples are in Appendix~\ref{sec: Appendix Human Evaluation details}.

\begin{figure}[t!]
\centering
\includegraphics[width=0.85\columnwidth]{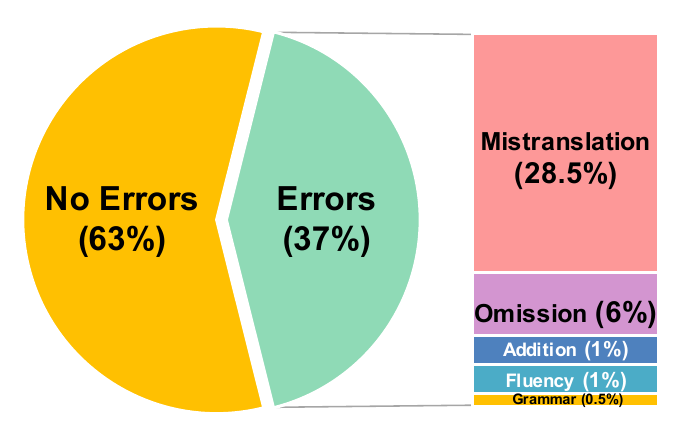}
\vspace{-2mm}
\caption{A distribution of different translation errors in sampled questions from Korean translate-test set.}
\vspace{-3mm}
\label{fig:annotation_summary}
\end{figure}
\begin{figure}[t!]
\centering
\includegraphics[width=0.9\columnwidth]{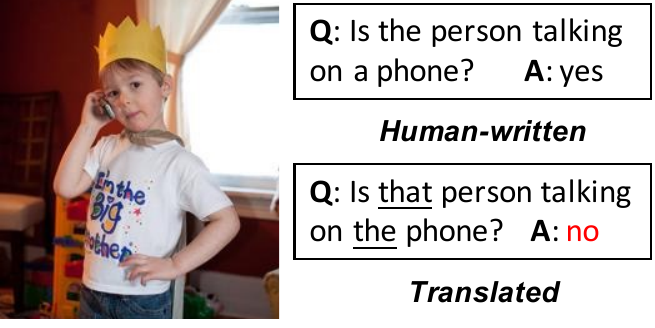}
\caption{A model is accurate for the original human-written question, but fails for a translated one. The Original Korean question is \textit{``그 사람이 전화 통화를 하고 있습니까?''}. Further annotation results are in Fig.~\ref{fig:case_study_results}.}
\vspace{-4mm}
\label{fig:case}
\end{figure}

As shown in Fig.~\ref{fig:annotation_summary}, although the model trained on human texts changes its prediction from the correct to wrong ones, a majority of translated questions (>60\%) do not contain crucial translation errors. 
In terms of translated questions without translation errors, most of them can be regarded as paraphrased sentences of their paired English questions as shown in Fig.~\ref{fig:case}. 
Based on these results, we confirm that models trained on human texts often make wrong predictions about translations that convey similar meanings to human ones. 
In other words, subtle differences between human and translated texts caused by translation artifacts indeed influence model behavior.

\subsection{Are Translation Artifacts Actually Presented in Translated Questions?}
\label{secesc:detector}

To enhance our comprehension of the increased translate-test accuracy of models trained on MT texts, we scrutinize model performances across samples categorized by the prevalence of translation artifacts.
Specifically, we quantify the \textit{human-likeness} of each translated question and assess its influence on VQA models.
To derive the human-likeness score $p_h(x)$ for every translated question $x$, we train a classifier based on RoBERTa~\citep{liu2019roberta}, designed to discern whether a given English question is written by a human or generated by round-trip (RT) translation.\footnote{The accuracy of the trained classifier is 86.97 in a class-balanced validation set.} 
After training, the classifier assigns a score $p_h(x)$ on how likely humans write each translate-test evaluation sample. 
We then categorize the translated evaluation samples into two groups with the same size - \textit{human-like} and \textit{NMT-like} - based on their respective human-likeness scores $p_h(x)$. 
The accuracy of VL-BERT models trained on different data origins~(\textit{i.e.}, human and NMT) is compared in these groups. 
More experimental details are in Appendix~\ref{sec: Appendix human likeness}.

\begin{table}[t!]
\small
\centering
\begin{adjustbox}{width=0.4\textwidth}
\begin{tabular}{c|cc}

\bottomrule
\backslashbox[20mm]{Train}{Test}& NMT-like & Human-like \\ \hline
Human  & 48.60 & 53.56 \\
NMT & 51.98 & 53.85 \\
\toprule

\end{tabular}
\end{adjustbox}
\caption{Averaged translate-test accuracy of VL-BERT models trained on different data origins.
Each column denotes the group of translate-test evaluation samples.}
\label{tab:human_like_accuracy}
\end{table}




From the results in Table~\ref{tab:human_like_accuracy}, we find that the model trained on human texts performs worse when the input questions are less likely to be written by humans. Specifically, the average accuracy across different target languages is 53.56 for human-like questions but 48.60 for NMT-like questions. 
Conversely, a model trained on MT texts shows less accuracy degradation on NMT-like questions compared to the one trained on human texts. 
These results indicate that VQA models are prone to make more errors when the given question is not likely to be written by humans, and training on (RT-) translated texts can alleviate such problems.

\begin{table}[t!]
\small
\centering
\vspace{2mm}
\begin{adjustbox}{width=0.4\textwidth}
\begin{tabular}{c|cc}
\bottomrule
\backslashbox[20mm]{Metric}{Test} & NMT-like & Human-like \\ \hline
TTR&92.52 & 95.14 \\
LD& 48.44&49.76 \\                
\toprule
\end{tabular}
\end{adjustbox}
\caption{Lexical diversity results of translate-test evaluation samples. TTR and LD denote the token-type ratio and lexical density, respectively.}
\vspace{-2mm}
\label{tab:lexical_analysis}
\end{table}


We next delve into the lexical diversity within each question group, inspired by previous findings that the translated texts are often simpler than human ones~\citep{zhang-toral-2019-effect, CLS_translationese}. 
Specifically, we use two metrics to measure the lexical diversity of translated questions used in the translate-test approach: (1) \textit{Token Type Ratio (TTR)} calculates the ratio of unique words over all words in the sentence, (2) \textit{Lexical Density (LD)} calculates the ratio of content words (words that likely to convey significant meaning - nouns, verbs, adverbs, and adjectives) over all words in the sentence. Lexical diversity results in Table~\ref{tab:lexical_analysis} indicate that NMT-like questions generally exhibit less variety in word usage. 
We suspect that such characteristics in translated questions make a difference in training and evaluation, resulting in performance degradation of models.

\begin{figure}[t!]
\centering
\includegraphics[width=0.45\textwidth]
{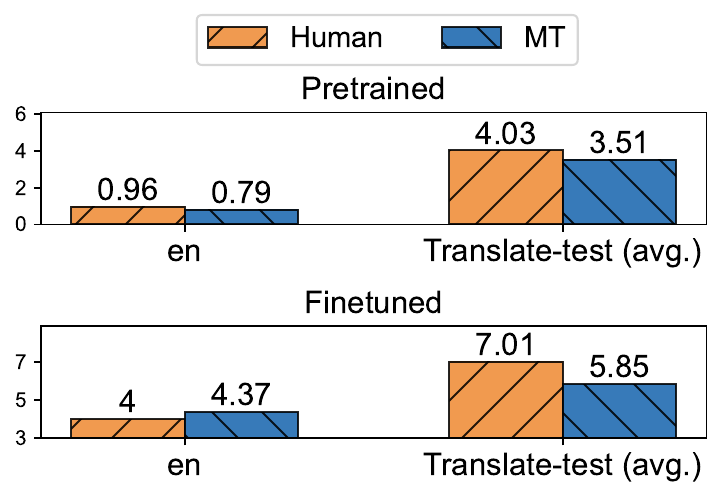}
\vspace{-3mm}
\caption{Representation discrepancy of translate-test evaluation samples against training samples from different data origins (Human and MT). 
Pretrained or finetuned VisualBERT is used to encode representation, and FID is used as a distance metric. 
A lower score indicates a low distance between training and evaluation samples. Full results across different languages are in Fig.~\ref{fig:fid}.}
\label{fig:fid_avg}
\end{figure}

\begin{table*}[t!]
\tiny
\centering
\begin{adjustbox}{width=1.0\textwidth}
\begin{tabular}{cccccccccc}
\toprule
\multirow{2}{*}{\textbf{\begin{tabular}[c]{@{}c@{}}RT\\ 
Pivot\end{tabular}}} & \multicolumn{1}{l}{\textbf{}} & \multicolumn{8}{c}{\textit{Translate-Test}} \\ \noalign{\vskip 0.5ex} \cline{3-10} 
& en  & bn & de & id & ko & pt & ru & zh & \textbf{avg.} \\ \hline
bn &
  53.93 &
  52.46 &
  53.18 &
  52.34 &
  50.93 &
  52.82 &
  52.54 &
  50.31 &
  52.08 \\
de &
  55.13 &
  \textbf{52.47} &
  \textbf{54.77} &
  52.68 &
  51.05 &
  54.10 &
  53.31 &
  49.72 &
  52.59 \\
id &
  54.64 &
  52.45 &
  53.53 &
  \textbf{53.57} &
  51.42 &
  53.60 &
  53.11 &
  \textbf{50.50} &
  52.60 \\
ko &
  53.62 &
  51.63 &
  52.73 &
  51.97 &
  \textbf{51.78} &
  52.72 &
  52.02 &
  50.41 &
  51.89 \\
pt &
  \textbf{55.64} &
  52.45 &
  54.53 &
  52.82 &
  50.96 &
  \textbf{55.02} &
  53.57 &
  49.52 &
  \textbf{52.69} \\
ru &
  54.94 &
  52.42 &
  54.10 &
  52.98 &
  51.15 &
  53.87 &
  \textbf{54.00} &
  50.31 &
  \textbf{52.69} \\
zh &
  51.56 &
  48.64 &
  49.55 &
  49.00 &
  48.20 &
  49.46 &
  48.88 &
  48.43 &
  48.88 \\
\toprule
\end{tabular}
\end{adjustbox}
\vspace{-2mm}
\caption{Evaluation results of models trained on RT translation with different pivot languages. Each row indicates the pivot language used in RT translation, and scores of all models with the same pivot languages are averaged. 
The highest scores in each column are highlighted in \textbf{bold}. 
Full results of all pivot languages and models are in Table~\ref{tab:pivot_full}.}
\vspace{-2mm}
\label{tab:varied_pivot}
\end{table*}

\subsection{Representation Analysis}
\label{secsec:representation_analysis}
Our previous observations reveal that translated texts exhibit distinct impacts compared to human ones when they are used for training and evaluation. 
We next analyze whether these different characteristics of translated texts appear in model representation. 
Specifically, we compare the representations of training samples from different origins (human and MT) against evaluation samples. 
As evaluation samples, we use the translate-test samples from different target languages and English evaluation samples written by human or RT translations.
We employ the penultimate layer output of visualBERT as the sample representation, and the Fréchet Inception Distance (FID)~\citep{fid} score is used to quantify the representation distance between training and evaluation samples. 
Additionally, to assess the impact of finetuning on model representation, we analyze VisualBERT at checkpoints before and after finetuning.

As shown in Fig.~\ref{fig:fid_avg}, we observe clear trends indicating that translated samples cluster more closely in the model representation space. 
In detail, all translate-test samples show lower FID scores with MT training samples than human ones. 
Note that these trends are consistent for both pretrained and finetuned models. 
These results indicate that characteristics shared within translated texts also affect the internal representation of VL models.

\begin{table}[t!]
\centering
\begin{adjustbox}{width=0.48\textwidth}
\begin{tabular}{l|ccccc}
\bottomrule
       \backslashbox[20mm]{Train}{Test}  & GMT   & \begin{tabular}[c]{@{}c@{}}M2M-\\ small\end{tabular} & \begin{tabular}[c]{@{}c@{}}M2M-\\ large\end{tabular} & \begin{tabular}[c]{@{}c@{}}NLLB-\\ small\end{tabular} & \begin{tabular}[c]{@{}c@{}}NLLB-\\ large\end{tabular} \\ \hline
Human  & 51.92 & 45.64 & 47.79 & 49.53 & 50.62 \\        
M2M-S  & 51.25 & 48.85 & 49.39 & 50.06 & 50.55 \\
M2M-L  & 52.31 & \textbf{49.37} & \textbf{50.43} & 50.94 & 51.48 \\
NLLB-S & 52.75 & 48.73 & 50.00 & 51.39 & 52.04 \\
NLLB-L & \textbf{53.18} & 48.59 & 50.04 & \textbf{51.65} & \textbf{52.52} \\ \toprule
\end{tabular}
\end{adjustbox}
\caption{Translate-test evaluation results with different MT systems to make RT-translated training and translate-test evaluation examples. 
Each row and column denote the origin of training and evaluation datasets, respectively.
The best scores on each evaluation set are highlighted in \textbf{bold}. 
Each score denotes the averaged accuracy of models described in Section~\ref{secsec:model}. Full results across languages and models are in Table~\ref{tab:tab:varying_nmt_full}.}
\vspace{-5mm}
\label{tab:varying_nmt}
\end{table}

\subsection{Varying NMT and Pivot Languages}
\label{secsec:vaied_nmt_and_pivot}

Based on our previous results, we confirm that addressing the misalignment of data origins between training and evaluation is effective for the translate-test approach. 
We now aim to understand how these benefits vary with changes in the MT systems or translation setups. To this end, we conduct experiments by varying (1) the MT system used for translating the training and evaluation sets and (2) the pivot language during the RT translation.

\noindent \textbf{Varied MT systems}
We use the following four MT systems in our experiments: \textsc{m2m-100-418m/1.2b}~\citep{M2M-100} and \textsc{nllb-200-600m/-3.3b}~\citep{NLLB}. 
Each MT system is used to make RT-translated training and translate-test evaluation sets. 
In detail, we use RT translation with different MT systems to make training sets, and the pivot language is fixed to German (de). 
All models described in Section~\ref{secsec:model} are individually trained on these four RT-translated datasets. 
For the evaluation set, we translate every target language into English using different MT systems, resulting in four different evaluation sets. 

Evaluation results are shown in Table~\ref{tab:varying_nmt}. 
Notably, models trained on translated texts usually outperform those trained on human texts in translate-test sets. These results suggest that, despite a mismatch between the MT systems used for RT translation and the translate-test, leveraging RT translation for training remains advantageous for cross-lingual transfer. 
In terms of MT system comparison, models usually show higher accuracy when MT systems used to make training and evaluation sets are in the same model family.
In the original English evaluation set, models with human texts perform best, followed by the ones with \textsc{nllb-200-3.3b} texts.

\begin{table}[t!]
\small
\centering
\begin{tabular}{l|ccc}
\toprule
\multicolumn{1}{c}{MT} & SacreBLEU & chrF & METEOR \\
\hline
\multicolumn{4}{c}{de $\rightarrow$ en} \\
\hline
M2M-Small & 30.82 & 55.55 & 0.61 \\
M2M-Large & 33.68 & 57.86 & 0.63 \\
NLLB-Small & 39.34 & 61.48 & 0.68 \\
NLLB-Large & \textbf{42.98} & \textbf{64.09} & \textbf{0.70} \\
\hline
\multicolumn{4}{c}{en $\rightarrow$ de} \\
\hline
M2M-Small & 25.77 & 54.40 & 0.55 \\
M2M-Large & 30.36 & 58.09 & 0.58 \\
NLLB-Small & 32.03 & 58.64 & 0.59 \\
NLLB-Large & \textbf{34.79} & \textbf{60.92} & \textbf{0.62} \\
\toprule
\end{tabular}
\caption{Evaluation results of different MT systems on IWSLT 2017 benchmarks~\citep{iwslt2017}. The best scores on each metric are highlighted in \textbf{bold}.}
\label{tab:mt_eval_result}
\end{table}
\begin{figure*}[t!]
\centering
\includegraphics[width=0.93\textwidth]{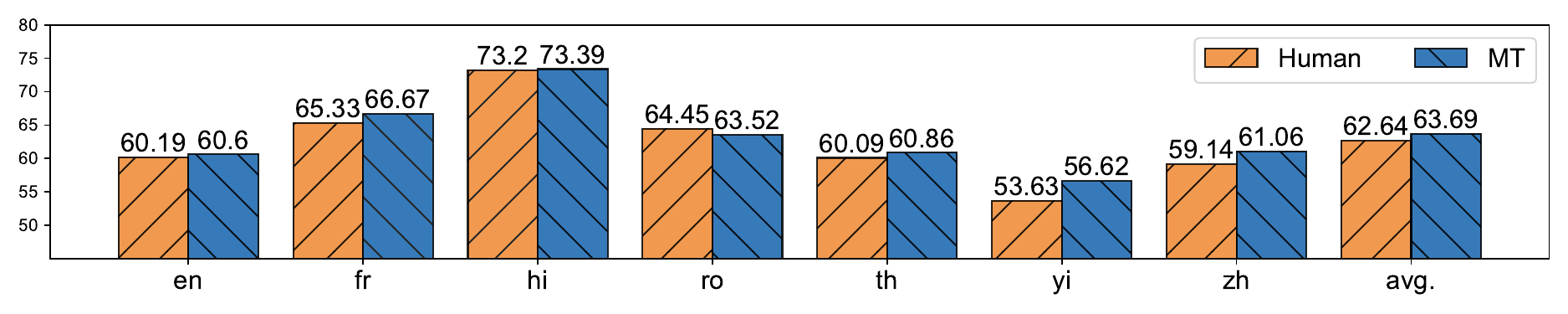}
\vspace{-3mm}
\caption{The averaged translate-test evaluation results of models with different training data origins on the yes/no question type in MaXM benchmark. 
The full results are presented in Table~\ref{tab:maxm_full}.}
\vspace{-1mm}
\label{fig:maxm}
\end{figure*}
\begin{table*}[t!]
\centering
\tiny
\begin{adjustbox}{width=0.93\textwidth}
\begin{tabular}{cccccccccc}
\toprule
 & \multicolumn{1}{l}{\textbf{}} & \multicolumn{8}{c}{\textit{Translate-Test}} \\ \noalign{\vskip 0.5ex} \cline{3-10} 
& en  & bn & de & id & ko & pt & ru & zh & \textbf{avg.} \\ \hline
Human & \textbf{56.91} & 51.01 & 54.10 & 52.17 & 50.58 & 54.02 & 50.84 & 49.99 & 51.82 \\
MT & 55.13 & 52.11 & 54.80 & 53.06 & 52.47 & 54.10 & 53.28 & 52.19 & 53.14 \\
MERGE & 56.52 & \textbf{52.80} & \textbf{55.54} & \textbf{53.75} & 53.04 & \textbf{55.08} & \textbf{54.10} & \textbf{52.73} & \textbf{53.86} \\
TAG & 56.67 & 52.65 & 55.44 & 53.56 & \textbf{53.11} & 54.83 & 53.89 & 52.65 & 53.73 \\
\toprule
\end{tabular}
\end{adjustbox}
\vspace{-2mm}
\caption{Data augmentation results. The highest scores in each column are highlighted in \textbf{bold}. All model scores with the same data origin are averaged. Full results are in Table~\ref{tab:da_full}.}
\label{tab:da_main}
\end{table*}

\subsection{Translation Quality of MT systems}
\label{sec: Appendix different MT systems Evaluation details}
From results in Section~\ref{secsec:vaied_nmt_and_pivot}, we observe that the accuracy ranking in the original English set and the GMT translate-test set is \textsc{nllb-200-3.3b} > \textsc{nllb-200-600m} > \textsc{m2m-100-1.2b} > \textsc{m2m-100-418m}. 
We suspect that this trend reflects the translation quality of the training data produced by each MT system. 
To corroborate this, we assessed these MT systems using an IWSLT 2017~\citep{iwslt2017} benchmark, while maintaining the same translation direction as in RT translation (i.e., en$\rightarrow$bn and vice versa). 
The IWSLT2017 dataset contains 8079 parallel sentences in these language directions, which involves multilingual text translation of TED talks. 
For evaluation, we utilized METEOR~\citep{banerjee-lavie-2005-meteor}, chrF~\citep{chrf}, and SacreBLEU~\citep{sacrebleu} as evaluation metrics.  
As shown in Table~\ref{tab:mt_eval_result}, we observe the results which are clearly aligned with the previously observed trends. 
Further details of each metric are in Table~\ref{tab:mt_metrics_eval}.

\noindent \textbf{Varied Pivot Language in RT Translation}
We vary the pivot languages used in RT translation to make different versions of translated training sets. All target languages presented in xGQA datasets are selected as pivot languages. As an MT system, we use \textsc{nllb-200-3.3b} to translate both training and evaluation samples. As shown in Table~\ref{tab:varied_pivot}, models usually show higher accuracy when a pivot language matches its target language. 
This tendency is consistent with previous findings~\citep{original_or_transalted}, where the texts within the same translation direction contain shared characteristics.

\subsection{Experiments with MaXM dataset}
\label{secsec:maxm}
We evaluate models trained on the xGQA dataset with MaXM~\citep{maxm}, a recently proposed evaluation-only benchmark for multilingual VQA. The MaXM dataset covers seven different languages: English~(en), French~(fr), Hindi~(hi), Hebrew~(iw), Romanian~(ro), Thai~(th), and Chinese~(zh). Each evaluation sample consists of an image, a question, and an answer. 
As the answers in the MaXM dataset are not exactly matched with the ones in xGQA that models are trained, we only use a question whose answer is either  ``yes'' or ``no''. 
More details about the MaXM dataset and full evaluation results are in Appendix~\ref{sec: Appendix xGQA details} and Table~\ref{tab:maxm_full}, respectively. 
As shown in Fig.~\ref{fig:maxm}, we observe results consistent with the xGQA dataset. Training on RT-translated texts increases the translate-test scores except for Romanian (ro) cases. 
\section{Reducing the Effect of Translation Artifact on Cross-lingual VQA}
\label{sec:DA}

Our findings demonstrate that training VQA models on translated texts induces higher accuracy in language transfer through the translate-test approach. 
Despite such gains, translated texts inevitably contain wrongly translated information due to the imperfection of MT systems. 
Moreover, as translations are known to be different from the naturally written human texts~\citep{volansky2015features,zhang-toral-2019-effect}, training models solely on the translated texts may degrade overall performance. These problems can be observed in our previous results; in Table~\ref{tab:translate-test_main}, the models trained on translated texts show a relatively low average score in the English evaluation set compared to those trained on human texts~(56.91$\rightarrow$55.13).

To resolve this, we leverage a simple data augmentation technique that uses both RT-translated texts and the original human-written texts for model training~(\textbf{MERGE}). Furthermore, following \citet{marie2020tagged}, we also adopt the approach that includes special tagging tokens in front of translated texts in both training and evaluation phases~(\textbf{TAG}). As MERGE and TAG double the number of training examples, we reduce the total training steps to half for a fair comparison across methods. 
Results with data augmentation methods are in Table~\ref{tab:da_main}. The accuracy of the original English evaluation set is increased in both MERGE and TAG compared to solely using translated samples. The overall scores for the translate-test are also improved with data augmentation. These results indicate that augmenting training data with both human and MT texts is helpful for cross-lingual transfer while maintaining its performance on the original English texts.

\section{Conclusion}
In this work, we analyze the impacts of translation artifacts presented in machine-translated English texts for cross-lingual VQA. 
Through extensive experiments, we find that current VL models usually suffer from distributional shifts caused by translation artifacts during cross-lingual transfer, resulting in undesirable performance degradation. 
As a remedy, we conduct experiments with simple data augmentation strategies and observe consistent performance gains. 

Our work focuses on translations that are semantically similar but written differently from human texts. 
In future work, we will explore mistranslation arising from context-free translation, where image information is not considered during a translation process.
To this end, recently advanced multimodal translation systems can be utilized~\citep{yao-wan-2020-multimodal}. 
Other important directions include considering a variance among different translations generated from a single text and devising an advanced training strategy to consider translation artifacts.

\section*{Limitations}
Our study is mainly conducted on a translate-test approach for a cross-lingual VQA task. 
We recognize that some of our results may not generalize other tasks, like image captioning. 
Nevertheless, as reasoning over natural language and image is a crucial ability for vision-language models, we believe it is a fundamental step to comprehend the impacts of translation in the VQA task to transfer across different languages seamlessly. 
Besides, since we mainly consider the conventional \textit{finetune-then-evaluate} pipelines, some experimental setups do not directly apply to recent models that do not perform parameter updates for learning (e.g., GPT-4V~\citep{OpenAI2023GPT4TR}).
As discussed in Appendix~\ref{sec: GPT4}, we observe that these models also can suffer from translation artifacts to some extent when performing VL tasks.
Performing extended analysis and proposals across diverse learning algorithms and models remains our future work.

\section*{Ethics Statement}
Most of the models in our experiments are trained on English datasets only, so the generalizability towards other source languages is not examined.
Besides, as the current MT systems are imperfect, training on translated texts may introduce unintended behaviors or favors to specific questions.
Future research should investigate such undesirable bias in translations and VQA models.

\section*{Acknowledgements}
This work was supported by a grant of the KAIST-KT joint research project through AI2X Laboratory, Tech Innovation Group, funded by KT (No. D23000019, Research on Multilingual Multicultural Vision-Language Representation Learning) and Institute for Information \& communications Technology Promotion(IITP) grant funded by the Korea government(MSIT) (No.RS-2019-II190075 Artificial Intelligence Graduate School Program(KAIST) and RS-2021-II212068, Artificial Intelligence Innovation Hub).

\bibliography{custom}

\newpage
\appendix
\section{Dataset Details}
\label{sec: Appendix xGQA details}
\paragraph{xGQA~\citep{xgqa}}
In our study, we used the English-balanced GQA~\citep{hudson2019gqa} training set for model training, which consists of 943k training examples and 72k training images. 
For model validation, the English GQA validation set, containing 132k samples and 10k images, is used. 
For evaluation, we used the balanced test-dev subset of the xGQA dataset, which includes 12,578 systematically structured questions with an average length of 8.5 words, associated with 398 images.  
The xGQA dataset extends the test-dev set of GQA by translating into seven different languages, each from a unique language family. 
In the translate-test approach, we used the official evaluation set released by~\citet{bugliarello-etal-2022-iglue}, which translates samples written in target languages into English with the Google Machine Translation system. 
Further details on the xGQA dataset are provided in~\citet{xgqa}.

\begin{table}[h!]
\centering
\begin{tabular}{lc}
\toprule
\textbf{Language} & \textbf{\# Examples} \\ \hline
English           & 75                   \\
French            & 70                   \\
Hindi             & 82                   \\
Hebrew            & 70                   \\
Romanian          & 77                   \\
Thai              & 75                   \\
Chinese           & 52  
\\ \toprule
\end{tabular}
\vspace{-2mm}
\caption{Number of selected examples for each language in MaXM dataset.}
\vspace{-2mm}
\label{tab:maxim statistics}
\end{table}

\paragraph{MaXM~\citep{maxm}}
The MAVERICS-XM3600 (MaXM) dataset, an evaluation-only VQA benchmark, originates from the Crossmodal-3600 dataset (XM3600)~\citep{crossmodal3600} and consists of translation-based question-answer pairs. 
MaXM includes 7 languages which are chosen based on their typological, genealogical, and geographical diversity. 
The statistics of selected evaluation samples for each language are presented in Table~\ref{tab:maxim statistics}.

\begin{table*}[t!]
\tiny
\centering
\begin{adjustbox}{width=0.9\textwidth}
\begin{tabular}{cccc}
\toprule
\multirow{2}{*}{Model}        & \multirow{2}{*}{Language Model}              & \multirow{2}{*}{Visual Tokens}                                                            & \multirow{2}{*}{\begin{tabular}[c]{@{}c@{}}\# Trainable Params (M) / \\ \# Total Params (M)\end{tabular}} \\
                              &                                              &                                                                                           &                                                                                                           \\ \hline

MUNITER                       & \href{https://huggingface.co/bert-base-multilingual-cased}{bert-base-multilingual-cased}                 & \begin{tabular}[c]{@{}c@{}}36 RoIs from Faster R-CNN \\ with ResNet-101\end{tabular}      & 116.46M / 116.46M                                                                                         \\ \hline
XUNITER                       & \href{https://huggingface.co/xlm-roberta-base}{xlm-roberta-base}                             & \begin{tabular}[c]{@{}c@{}}36 RoIs from Faster R-CNN \\ with ResNet-101\end{tabular}      & 116.46M / 116.46M                                                                                         \\ \hline
UC$^2$                           & \href{https://huggingface.co/xlm-roberta-base}{xlm-roberta-base}                             & \begin{tabular}[c]{@{}c@{}}36 RoIs from Faster R-CNN \\ with ResNet-101\end{tabular}      & 281.64M / 281.64M                                                                                         \\ \hline
M$^3$P                           & \href{https://huggingface.co/xlm-roberta-base}{xlm-roberta-base}                             & \begin{tabular}[c]{@{}c@{}}10-100 RoIs from Faster R-CNN \\ with ResNeXt-101\end{tabular} & 376.90M / 376.90M                                                                                                                 \\           \hline                    
LxMERT                        & \href{https://huggingface.co/bert-base-uncased}{bert-base-uncased}                                & \begin{tabular}[c]{@{}c@{}}36 RoIs from Faster R-CNN \\ with ResNet-101\end{tabular}      & 213.33M / 213.33M                                                                                         \\ \hline
UNITER                        & \href{https://huggingface.co/bert-base-uncased}{bert-base-uncased}                                & \begin{tabular}[c]{@{}c@{}}36 RoIs from Faster R-CNN \\ with ResNet-101\end{tabular}      & 116.46M / 116.46M                                                                                         \\ \hline
VILBERT                       & \href{https://huggingface.co/bert-base-uncased}{bert-base-uncased}                                & \begin{tabular}[c]{@{}c@{}}36 RoIs from Faster R-CNN \\ with ResNet-101\end{tabular}      & 244.04M / 244.04M                                                                                         \\ \hline
VisualBERT                    & \href{https://huggingface.co/bert-base-uncased}{bert-base-uncased}                                & \begin{tabular}[c]{@{}c@{}}36 RoIs from Faster R-CNN \\ with ResNet-101\end{tabular}      & 116.84M / 116.84M                                                                                         \\ \hline
VL-BERT                       & \href{https://huggingface.co/bert-base-uncased}{bert-base-uncased}                                & \begin{tabular}[c]{@{}c@{}}36 RoIs from Faster R-CNN \\ with ResNet-101\end{tabular}      & 118.03M / 118.03M                                                                                         \\ \hline

\multirow{2}{*}{BLIP-2}         & \multirow{2}{*}{\href{https://huggingface.co/facebook/opt-2.7b}{opt-2.7b}}                    & \multirow{2}{*}{-}                                                                        & \multirow{2}{*}{190.29M / 3827.78M}                                                                       \\
                              &                                              &                                                                                           &                                                                                                           \\ \hline
\multirow{2}{*}{InstructBLIP} & \multirow{2}{*}{\href{https://huggingface.co/google/flan-t5-xl}{flan-t5-xl}}                  & \multirow{2}{*}{-}                                                                        & \multirow{2}{*}{189.27M / 4024.92M}                                                                       \\
                              &                                              &                                                                                           &                                                                                                           \\ \hline
\multirow{2}{*}{FLAVA}        & \multirow{2}{*}{ViT-B/16 based text encoder} & \multirow{2}{*}{-}                                                                        & \multirow{2}{*}{243.36M /  243.36M}                                                                       \\
                              &                                              &                                                                                           &                                                                                                           \\ \hline
          
\toprule
\end{tabular}
\end{adjustbox}
\caption{We report the key properties, training parameters, and total parameters for all the models. 
}
\label{tab:details_models}
\end{table*}
\section{Model Details}
\label{sec: Appendix Model details}
Table~\ref{tab:details_models} summarizes the key characteristics of all models described in Section~\ref{secsec:model}. 
For visual tokens, we utilize 36 image regions from a ResNet101 backbone~\citep{resnet}, and 10 to 100 image regions from a ResNeXt-101 backbone~\citep{resnext}.
For BLIP-2, InstructBLIP, and FLAVA, we use the official implementations released by authors.
For other models, we use the implementation released by \citet{bugliarello-etal-2022-iglue}.

\section{Translation Details}
\label{sec: Appendix Translation}
\noindent \textbf{RT Translation} We use roundtrip~(RT) translation to make translated English training dataset.
Unless otherwise specified, \textsc{NLLB-200-3.3B} is used as an MT system, and German~(de) is used as a pivot language.
Following ~\citet{artetxe2023revisiting}, we use stochastic and deterministic decoding strategies for RT translation.
Specifically, for forward translation~(en $\rightarrow$ de), we use nucleus sampling~\citep{holtzman2019curious} with $p=0.9$.
For backward translation~(de $\rightarrow$ en), we use beam search with beam size as 5.
For both translation directions, the maximum number of repeated n-gram is set to 5.

\noindent \textbf{Translate-Test}
Unless otherwise specified, we use the evaluation set released by \citet{bugliarello-etal-2022-iglue} for a fair comparison.
When constructing the translate-test evaluation set ourselves, as in Section~\ref{secsec:vaied_nmt_and_pivot}, we use beam search with beam size 4.

\noindent \textbf{Translate-Train}
We translate the original English training set into every target language in xGQA.
\textsc{NLLB-200-3.3B} is used as an MT system for this process, and beam search is used with beam size 5.

\section{Implementation Details}
\label{sec: Appendix Implementation details}
For finetuning VL models, we follow hyperparameters described in \citet{bugliarello-etal-2022-iglue} for a fair comparison.
Specifically, we utilize the AdamW optimizer~\citep{adamw} with betas set at $(0.9, 0.999)$ and $\epsilon$=1e-8. 
The maximum number of tokens in the input sequence is set to 40, and the batch size is set to 256.
The total training epochs are set to 5.
The learning rate is set to 1e-4, and a linear learning late scheduler is used with a 0.5 warm-up epoch.
For training, we used cross-entropy loss for all 1,842 labels available in the GQA dataset.
In overall experiments, a single NVIDIA-A100 GPU with 40GB of memory is used for BLIP-2, InstructBLIP, and FLAVA, and a single model is trained in one day.
Other models are trained with a 3090 RTX GPU with 24GB of memory and are trained in 5 hours.
The experiments are implemented with PyTorch~\citep{paszke2019pytorch}.

\section{Statistical Test Results}
\label{sec: Appendix Statistical Test}
\begin{table}[t!]
\centering
\vspace{2mm}
\begin{tabular}{c|c}
\toprule
\textbf{Language}          & \textbf{p-value} \\ \hline
bn (RT \textgreater Human) & 2.89e-17        \\
de (RT \textgreater Human) & 3.51e-15       \\
id (RT \textgreater Human) & 2.48e-17        \\
ko (RT \textgreater Human) & 1.70e-26        \\
pt (RT \textgreater Human) & 7.33e-05       \\
ru (RT \textgreater Human) & 2.14e-14       \\
zh (RT \textgreater Human) & 5.21e-27    \\
\bottomrule

\end{tabular}
\caption{
We gather all the results of each row in Table~\ref{tab:translate-test_three_times_run} to get the results for each model and performed a t-test on these aggregated results. Here $\textit{RT} > 
\textit{Human}$ means models trained with round-trip translated texts (\textit{RT}) are better than models trained with human texts (\textit{Human}).}
\label{tab:significant langugage}
\end{table}
\begin{table}[t!]
\centering
\vspace{2mm}
\begin{adjustbox}{width=0.35\textwidth}
\begin{tabular}{l|c}
\toprule
\textbf{Language}          & \quad \textbf{p-value} \quad \\ \hline
MUNITER (RT > Human) & 3.78e-07         \\
XUNITER(RT > Human) & 1.26e-04       \\
UC$^2$ (RT > Human) &  2.06e-08       \\
M$^3$P(RT > Human) & 1.35e-07       \\
LXMERT (RT \textgreater Human) &  2.43e-06       \\
UNITER (RT \textgreater Human) & 1.05e-07    \\
VILBERT (RT \textgreater Human) & 8.29e-07     \\
VisualBERT (RT \textgreater Human) & 6.76e-07    \\
VL-BERT (RT \textgreater Human) & 1.56e-07     \\
BLIP-2 (RT \textgreater Human) & 4.82e-10     \\
InstructBLIP (RT \textgreater Human) &  5.14e-07       \\
FLAVA (RT \textgreater Human) & 1.62e-07     \\
\bottomrule
\end{tabular}
\end{adjustbox}
\caption{We gather all the results of each column in Table~\ref{tab:translate-test_three_times_run} to get the results for each language and perform a t-test. Here $\textit{RT} > 
\textit{Human}$ means models trained with round-trip translated texts (\textit{RT}) are better than models trained with human texts (\textit{Human}).}
\vspace{-2mm}
\label{tab:significant model}
\end{table}
Based on the findings in Section~\ref{secsec:results}, we observe an improvement in test accuracy when the data origins for training and evaluation are aligned.
To demonstrate that this improvement consistently occurs in the translate-test, we train models three times with different seeds and report the average performance of models in Table~\ref{tab:translate-test_three_times_run}.

Furthermore, we perform significance tests to demonstrate that this improvement consistently occurs in the translate-test.
Specifically, we conduct significant tests on all the aggregated results as well as on the language-specific and model-specific results in Table~\ref{tab:translate-test_three_times_run}.\footnote{For significant tests, we use the paired t-test with $\alpha=0.05$.} 
Our evaluation specifically compared the performance of models trained with MT texts against those trained with human texts in the translate-test.

First, we aggregate all the outcomes from models trained on machine-translated texts and compare them to those trained on human texts in Table~\ref{tab:translate-test_three_times_run}. 
The results demonstrate a significant advantage in training with machine-translated texts over human texts, with a p-value of 6.49e-66.

Moving on to specific language results in Table~\ref{tab:significant langugage}, we gather all the model outcomes for each language except for English. 
We then compare the performance of models trained on machine-translated texts to those trained on human texts.

When considering the performance of individual models, we collect the results from all evaluated languages for each model (\textit{i.e.}, collect all the results of each row in Table~\ref{tab:translate-test_three_times_run}). We then compare the performance of models trained on machine-translated texts to those trained on human texts in Table~\ref{tab:significant model}.

As a result, we can observe that evaluating models trained on human-written English data with translated texts could negatively impact the generalization of models to other languages. By simply aligning the data origins for both training and evaluation sets, the overall performance in the translate-test can be improved. 

\begin{figure}[t!]
\centering
\includegraphics[width=0.75\columnwidth]{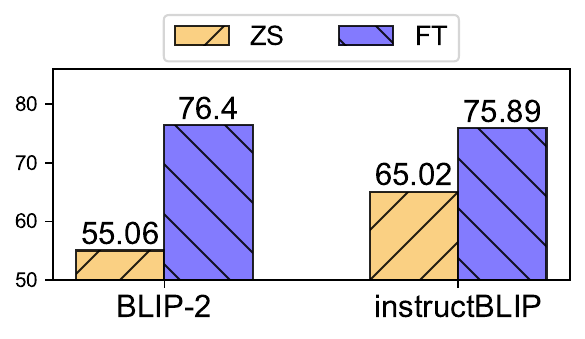}
\vspace{-3mm}
\caption{Comparison of zero-shot and finetuning with yes/no questions in xGQA. 
\textit{ZS} and \textit{FT} denote the accuracy of zero-shot and finetuned models, respectively.}
\label{fig:lm_zero_shot}
\end{figure}

\section{Zero-shot Evaluation of VL models}
\label{sec: zeroshot_eval}

Recent VL models like BLIP-2~\citep{BLIP2} or InstructBLIP~\citep{instructBLIP} can perform target downstream tasks without task-specific finetuning by relying on its language generation ability.
To compare the effectiveness of finetuning on these models, we evaluate the models with and without finetuning by using xGQA evaluation samples whose answers are either ``yes'' or ``no''.
This subset contains 4,525 samples out of 12,578 total evaluation samples.
For zero-shot evaluation, we prompt the model with task description as follows: `\texttt{Answer the following question in ``yes'' or ``no''.\textbackslash n Question: }{\textbf{<question>}} \texttt{\textbackslash n Answer: }'.
Models generate the next token as an answer for the given question in the prompt with an image.
We conduct post-processing steps including case-normalization or punctuation mark removal to derive the binary prediction of models.
For model implementation, we use the models released by \citet{wolf-etal-2020-transformers}.

As shown in Fig.~\ref{fig:lm_zero_shot}, although models exhibit competitive zero-shot scores, their performance is lower than finetuned ones. These results imply that finetuning the models on task-specific datasets is also crucial for recent VL models. In this regard, it is still essential for the VL models to consider and address data origin misalignment presented in training and evaluation.

\section{Human Evaluation Details}
\label{sec: Appendix Human Evaluation details}
We first identified examples where questions, initially correct in English, became incorrect in the translate-test. 
Among these examples, we specifically focused on cases where the UC$^2$ model, trained using the original English GQA dataset, provided incorrect results, but the UC$^2$ model trained with RT-translated data generated correct responses. 
From the examples that conformed to these restrictions, we analyzed a subset of 200 examples. 

Following~\citet{moghe-etal-2023-extrinsic}, we annotated any machine translation (MT) errors in these examples, utilizing the Multidimensional Quality Metrics (MQM) ontology~\citep{burchardt-2013-multidimensional}. 
This framework categorizes errors into a hierarchical structure, allowing for the evaluation of translations based on this hierarchy.  
Our analysis focused on 5 error types within the MQM ontology, including \textit{Mistranslation}, \textit{Addition}, \textit{Omission}, \textit{Fluency}, and \textit{Grammar}. 
Two authors with a master's degree or higher separately annotated each evaluation sample.
The annotated examples from our case study are presented in Fig.~\ref{fig:case_study_results}.

\section{Human-likeness Analysis Details}
\label{sec: Appendix human likeness}
We use a confidence score of a text classifier to analyze the prevalence of translation artifacts in every translated question. 
The classifier is trained to discriminate whether the data origin of a given question is a human or MT system.
Specifically, we finetune RoBERTa-base~\citep{liu2019roberta} to classify whether the given question is from the original human-written dataset or the translated dataset from another target language.
The training epochs, batch size, and learning rate are set to 3, 24, and 2e-5.
The finetuned classifier assigns the confidence score $p_h(x)$ about how likely a human writes the question to each translated question in translate-test evaluation sets.
This confidence score is regarded as \textit{human-likeness} of translated questions.

\begin{table*}[t!]
\centering
\begin{adjustbox}{width=0.9\textwidth}
\begin{tabular}{lll}
\toprule
\textbf{Method}            & \textbf{Code}                                                            & \textbf{Notes}                                                                           \\ \hline
\multirow{3}{*}{METEOR}    & \multirow{3}{*}{\url{https://huggingface.co/spaces/evaluate-metric/meteor}}    & \multirow{3}{*}{}                                                                        \\
                           &                                                                          &                                                                                          \\
                           &                                                                          &                                                                                          \\ \hline
\multirow{3}{*}{chrF}      & \multirow{3}{*}{\url{https://huggingface.co/spaces/evaluate-metric/chrf}}     & \multirow{3}{*}{\texttt{signature: ``nrefs:1|case:mixed|eff:no|tok:13a|smooth:exp|version:2.0.0''}} \\
                           &                                                                          &                                                                                          \\
                           &                                                                          &                                                                                          \\ \hline
\multirow{3}{*}{SacreBLEU} & \multirow{3}{*}{\url{https://huggingface.co/spaces/evaluate-metric/sacrebleu}} & \multirow{3}{*}{\texttt{signature: ``nrefs:1|case:mixed|eff:no|tok:13a|smooth:exp|version:2.0.0''}} \\
                           &                                                                          &                                                                                          \\
                           &                                                                          &                                                                          
                           \\
                           \toprule
\end{tabular}
\end{adjustbox}

\caption{Code and versions for each MT metric.}
\label{tab:mt_metrics_eval}
\end{table*}

\begin{figure}[t!]
\centering
\includegraphics[width=0.95\columnwidth]{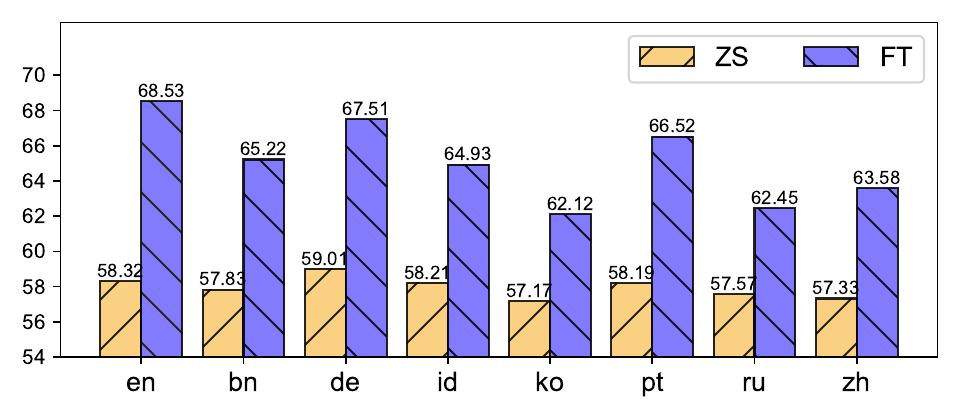}
\caption{Accuracy of LLaMA-Adapter-V2 zero-shot and fine-tuned performance on yes/no questions on xGQA. 
The finetuned model is trained on all human training samples in the xGQA dataset.}
\label{fig:llama_yesno}
\end{figure}
\begin{table*}[t!]
\centering
\tiny
\begin{adjustbox}{width=0.9\textwidth}
\begin{tabular}{cccccccccc}
\bottomrule
\multicolumn{1}{l}{} & en & bn & de & id & ko & pt & ru & zh & \textbf{avg.} \\
\hline
Zero-Shot & 60.67 & 64.67 & 65.33 & 61.67 & 66.00 & 64.33 & 62.33 & 59.67 & 63.43 \\
Translate-Test & - & 57.33 & 60.67 & 59.00 & 56.33 & 59.00 & 55.33 & 60.33 & 58.28 \\ \toprule
\end{tabular}
\end{adjustbox}
\caption{
Evaluation results of \url{gpt-4-1106-vision-preview} on xGQA datasets. All experiments are conducted based on 300 yes/no type questions. \textit{Zero-Shot} denotes that the input question is written in the target language.
}
\label{tab:gpt4_quantitative}
\end{table*}
\begin{table*}[t!]
\tiny
\centering
\begin{adjustbox}{width=0.95\textwidth}
\begin{tabular}{cccccccccc}
\toprule 
  & \multicolumn{1}{l}{\textbf{}} & \multicolumn{8}{c}{\textit{Translate-Test}} \\ \noalign{\vskip 0.5ex} \cline{3-10} 
 & en  & bn & de & id & ko & pt & ru & zh & \textbf{avg.} \\ \hline
Human   & 53.03& 47.72& 50.40& 48.10& 46.60& 50.08& 48.01& 46.90& 48.26 \\
MT      & 51.41& 48.76& 51.11& 49.36& 48.93& 50.64& 49.98& 48.94& 49.67 \\
MERGE   & 53.15& \textbf{49.52}& \textbf{51.79}& \textbf{50.25}& \textbf{49.73}& 51.64& \textbf{50.57}& \textbf{49.42}& \textbf{50.42} \\
TAG     & \textbf{53.22}& 49.21& 51.74& 50.23& 49.54& \textbf{52.04}& 50.38& 49.32& 50.35 \\
\toprule
\end{tabular}
\end{adjustbox}

\caption{Evaluation results of LLaMA-Adapter-V2 models parameter-efficiently finetuned with different data origins. The highest scores in each column are highlighted in \textbf{bold}.}
\vspace{-4mm}
\label{tab:llama_main}
\end{table*}

\section{Experiments with LLaMA-Adapter-V2}
\label{sec: Appendix LLAMA}

We examine whether LLaMA-Adapter-V2~\citep{llama_adapter_v2}, a recently proposed powerful VL model with a large language model, also suffers from translation artifacts for cross-lingual VQA tasks. 
To this end, we finetune LLaMA-Adapter-V2 with different training options~(Human, MT, MERGE, and TAG) and observe their results. 
Specifically, we add a classification head on top of end-of-sequence~(eos) token representation in LLaMA and finetune it along with the unfrozen weights.\footnote{\url{LORA-BIAS-7B} is used.}
The model is parameter-efficiently finetuned, where only a small portion of the total parameters are updated~(15M). 
We use the official codes released by the authors\footnote{\url{https://github.com/OpenGVLab/LLaMA-Adapter/tree/main/llama_adapter_v2_multimodal7b}} for implementation, and LLaMA-7b~\citep{touvron2023llama} with CLIP visual encoder~\citep{CLIP} is used.
The overall finetuning setups follow previously mentioned ones in Section~\ref{secsecsec: implementation details}.
Note that we also finetune and evaluate models to directly generate the answer text, but the scores are usually lower compared to using the classification head.

As shown in Table~\ref{tab:llama_main}, leveraging translated texts for training is beneficial to the translate-test approach of LLaMA-Adapter-V2, where the models trained on translated texts show higher accuracy compared to human texts. 
MERGE and TAG further improve accuracy in English and other target languages.

Besides, we also evaluate the model without finetuning on xGQA to probe its zero-shot ability. 
Since zero-shot classification with generation models requires roughly the number of forward passes with answer candidates, we choose evaluation samples whose label is either “yes” or “no”, and measure the probability of both tokens. 
Regarding the comparison with zero-shot and finetuning for yes/no question types in Fig.~\ref{fig:llama_yesno}, the finetuned model scores better than the zero-shot approach. 
This result implies that although recent VL models show impressive zero-shot capability, finetuning on task-specific datasets is still required for better performance. 

\section{Experiments with GPT-4-Vision}
\label{sec: GPT4}
In this study, we present experimental results of GPT-4-Vision~\citep{OpenAI2023GPT4TR}, a cutting-edge VL model. 
We use 300 evaluation samples of yes/no questions described in Appendix~\ref{sec: zeroshot_eval}.
We include all target languages and their corresponding original English questions. 
For evaluations in the target languages, inputs consist of questions either originally written in the target language or translated into English via GMT. 
The prompt format and the evaluation outcomes are presented in Fig.~\ref{tab:gpt4_vision}, and Table~\ref{tab:gpt4_quantitative}, respectively.

Our findings indicate that GPT-4 can serve as an effective multilingual VL model. 
Remarkably, its performance in all languages except Chinese exceeds that of English.
Directly using the target language proves more efficient than relying on the translated source language, primarily due to the inherent errors in translation processes.

However, GPT-4 falls short of the finetuned monolingual models detailed in Appendix~\ref{sec: zeroshot_eval}. 
The direct comparison between GPT-4 and these models is nuanced, largely because of differences in evaluation settings.\footnote{This complexity arises from the differences in the number of questions asked and the categorization of any response from GPT-4 other than ``yes'' or ``no'' as incorrect.}
Despite these challenges, the translate-test with strong VL models yielded more favorable outcomes than using GPT-4, with scores of 63.43 compared to 76.4 and 75.89 for finetuned BLIP-2 and InstructBLIP models, respectively,
Additionally, our qualitative analysis indicates that GPT-4 is also susceptible to translation artifacts, which can cause differences in predictions between human and MT texts.
We present the qualitative results of GPT-4 on xGQA in Fig.~\ref{tab:gpt4_vision} and \ref{tab:gpt4_vision_continued}.

\begin{figure*}[t!]
\centering
\includegraphics[width=0.9\textwidth]{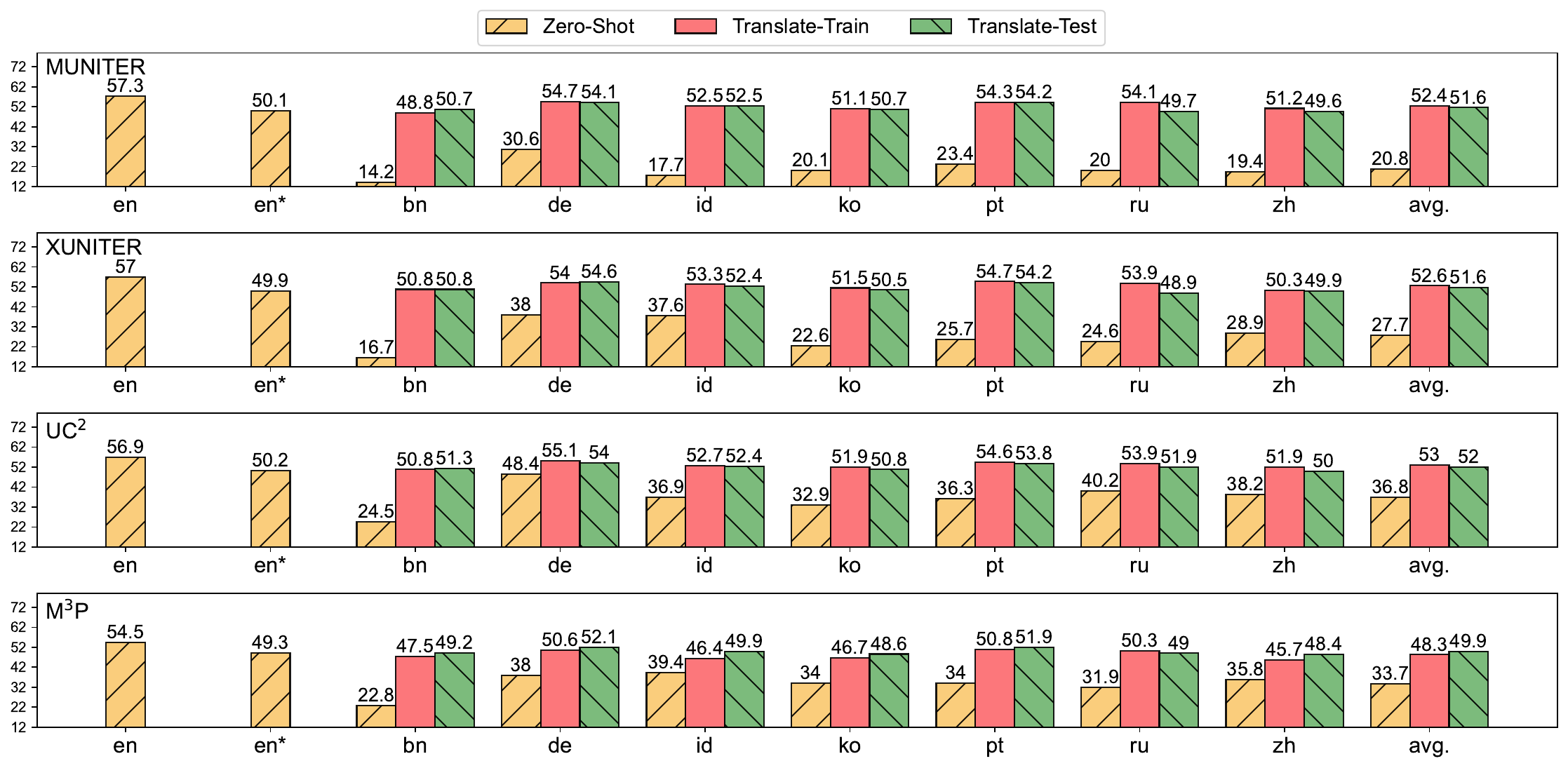}
\vspace{-1mm}
\caption{\textbf{Multilingual models results} The indicators are the same as Fig.~\ref{fig:multilingual_avg}.}
\label{fig:multilingual_full}
\end{figure*}

\section{Experiments on Image Captioning}
We further probed whether the problematic issues raised by translation artifacts are also present in an image captioning task. 
To this end, we used a multilingual image captioning dataset, namely Crossmodal-3600~\citep{crossmodal3600}, and a BLIP-2-large finetuned on the COCO dataset~\citep{lin2014microsoft}.
In detail, we first generated English captions for all images in the test dataset and translated these generations into individual target languages. This generate-and-translate process is designed to resemble the translate-test approach of cross-lingual VQA, where only an English-tailored VL model and a machine translation (MT) system were available. We then evaluated the translated captions with a reference-based metric (i.e., CiDEr~\citep{vedantam2015cider}), and two types of answer captions in target languages were considered as a reference set: (1) Human-written reference captions presented in Crossmodal-3600, and (2) RT-translated reference captions created by RT-translation (i.e., target language->en->target language) of the original reference captions. Using CiDEr, we calculated the similarity score between generated and reference captions.

As shown in Table~\ref{tab:captioning}, we observed clear trends that the generated captions achieve higher scores with RT-translated reference captions, which confirms that issues of cross-lingual multimodal tasks derived from translation artifacts are not just limited to VQA tasks. Intuitively, if both generated and RT-translated captions - translated texts - exhibit shared characteristics that deviate from human-written texts (i.e., translation artifacts), generated captions are likely to receive higher scores with RT-translated references than human-written ones. We leave the impacts of our data augmentation strategies on these generative multimodal tasks as our future work.

\begin{table*}[t!]
\centering
\tiny
\begin{adjustbox}{width=0.9\textwidth}
\begin{tabular}{cccccccc}
\bottomrule
\multicolumn{1}{l}{} & bn & de & id & ko & pt & ru & \textbf{avg.} \\
\hline
Human-ref & 0.159&	0.298&	0.382	&0.114	&0.400	&0.230	&0.26 \\
Translated-ref&0.344&	0.381	&0.554&	0.219&	0.471&	0.282&	0.375 \\
\toprule
\end{tabular}
\end{adjustbox}
\caption{
Translate-test results of an image captioning task with different types of reference captions. The \textit{Human-ref} and \textit{Translated-ref} denote the original human-written target-language references and their RT-translated references, respectively.
}
\label{tab:captioning}
\end{table*}

\section{Additional Results}
\noindent \textbf{Full results of Fig.~\ref{fig:multilingual_avg}}
Fig.~\ref{fig:multilingual_full} presents full results of Fig.~\ref{fig:multilingual_avg}.


\begin{table*}[t!]
    \tiny
    \centering
    
    \begin{adjustbox}{width=0.98\textwidth}
        \begin{tabular}{ccc|cccccccc}
            \midrule
            \textbf{Models} & \multicolumn{1}{l}{\textbf{RT?}} & \multicolumn{1}{c}{en} & bn & de & id & ko & pt & ru & zh & \textbf{avg.}\\ \noalign{\vskip 0.15ex}
            \hline         
             &  & \underline{57.20} & 50.79 & 53.95 & 52.40 & 50.60 & 54.30 & 49.85 & 49.63 & 51.65  \\
            \rowcolor[rgb]{0.882,0.882,0.882} \multirow{-2}{*}{MUNITER} \cellcolor[rgb]{1,1,1} &  \checkmark  & 55.57 & \underline{52.14} & \underline{55.36} & \underline{53.35} & \underline{52.92} & \underline{54.54} & \underline{53.68} & \underline{52.43} & \underline{53.49} \\ 
             
             &  & \underline{56.96} & 51.20 & 54.46 & 52.39 & 50.70 & \underline{54.19} & 49.81 & 49.92 & 51.81 \\
             \rowcolor[rgb]{0.882,0.882,0.882} \multirow{-2}{*}{XUNITER} \cellcolor[rgb]{1,1,1}&  \checkmark  &55.27 & \underline{52.06} & \underline{54.88} & \underline{52.76} & \underline{52.32} & 54.01 & \underline{53.01} & \underline{52.20} & \underline{53.03}\\
             
            &  & \underline{56.92} & 51.34 & 54.27 & 52.43 & 51.20 & 54.25 & 52.49 & 50.16 & 52.30 \\
             \rowcolor[rgb]{0.882,0.882,0.882} \multirow{-2}{*}{UC$^2$} \cellcolor[rgb]{1,1,1}&  \checkmark  & 55.50 & \underline{52.60} & \underline{55.29} & \underline{53.54} & \underline{53.27} & \underline{54.52} & \underline{53.96} & \underline{52.82} & \underline{53.71}\\
             
             &  & \underline{54.70} & 48.65 & 51.43 & 49.59 & 47.94 & 51.37 & 48.35 & 47.99 & 49.33 \\
              \rowcolor[rgb]{0.882,0.882,0.882} \multirow{-2}{*}{M$^3$P}\cellcolor[rgb]{1,1,1}&  \checkmark  & 53.51 & \underline{49.93} & \underline{52.51} & \underline{50.72} & \underline{49.97} & \underline{51.77} & \underline{50.97} & \underline{49.97} & \underline{50.84}\\
             
             &  & \underline{54.89} & 48.94 & 52.39 & 50.36 & 48.58 & 52.20 & 47.59 & 47.76 & 49.69 \\
             \rowcolor[rgb]{0.882,0.882,0.882} \multirow{-2}{*}{LXMERT} \cellcolor[rgb]{1,1,1}&  \checkmark  & 53.66 & \underline{50.65} & \underline{53.09} & \underline{51.59} & \underline{50.83} & \underline{52.59} & \underline{51.65} & \underline{50.48} & \underline{51.55}\\
             
             &  & \underline{57.52} & 51.44 & 54.37 & 52.64 & 51.21 & 54.45 & 51.86 & 50.22 & 52.32 \\
             \rowcolor[rgb]{0.882,0.882,0.882}\multirow{-2}{*}{UNITER}\cellcolor[rgb]{1,1,1}&  \checkmark  & 55.92 & \underline{52.34} & \underline{55.45} & \underline{53.60} & \underline{53.13} & \underline{54.75} & \underline{53.83} & \underline{52.67} & \underline{53.68}\\
             
            &   & \underline{57.08} & 51.05 & 54.37 & 52.68 & 51.04 & 54.20 & 50.14 & 50.01 & 51.93 \\
             \rowcolor[rgb]{0.882,0.882,0.882}\multirow{-2}{*}{VILBERT} \cellcolor[rgb]{1,1,1}&  \checkmark  & 54.84 & \underline{52.58} & \underline{55.19} & \underline{53.79} & \underline{52.97} & \underline{54.53} & \underline{53.92} & \underline{52.72} & \underline{53.67}\\
             
            &   & \underline{55.26} & 49.51 & 52.43 & 50.24 & 48.80 & \underline{52.54} & 50.62 & 48.62 & 50.40 \\
             \rowcolor[rgb]{0.882,0.882,0.882}\multirow{-2}{*}{VisualBERT} \cellcolor[rgb]{1,1,1}&  \checkmark  & 53.59 & \underline{50.53} & \underline{53.10} & \underline{51.07} & \underline{50.49} & 52.44 & \underline{51.41} & \underline{50.63} & \underline{51.38}\\
             
             &   & \underline{57.66} & 51.03 & 53.95 & 52.39 & 50.78 & 54.58 & 50.49 & 49.73 & 51.85 \\
             \rowcolor[rgb]{0.882,0.882,0.882}\multirow{-2}{*}{VL-BERT}\cellcolor[rgb]{1,1,1}&  \checkmark  & 55.67 & \underline{52.37} & \underline{55.27} & \underline{53.55} & \underline{52.81} & \underline{54.76} & \underline{53.54} & \underline{52.32} & \underline{53.52}\\
             
            &  & \underline{57.84} & 51.59 & 54.52 & 52.73 & 51.26 & 54.61 & 52.02 & 51.06 & 52.54 \\
             \rowcolor[rgb]{0.882,0.882,0.882}\multirow{-2}{*}{BLIP-2} \cellcolor[rgb]{1,1,1}&  \checkmark  & 56.35 & \underline{53.36} & \underline{55.86} & \underline{54.13} & \underline{53.54} & \underline{55.12} & \underline{54.42} & \underline{53.40} & \underline{54.26}\\

             &  & 
             \underline{57.76}&	51.65&	54.81&	53.04	&51.08	&54.79&	52.92&	51.53&	52.83 \\
             \rowcolor[rgb]{0.882,0.882,0.882}\multirow{-2}{*}{InstructBLIP} \cellcolor[rgb]{1,1,1}&  \checkmark  & 
             56.15 & \underline{53.20} & \underline{55.54} & \underline{53.90} & \underline{53.36} & \underline{54.74} & \underline{54.45} & \underline{53.19} & \underline{54.05} \\
             
             &  & \underline{\textbf{58.43}} & 53.27 & 55.90 & 54.00 & 52.66 & 55.72 & 53.32 & 51.73 & 53.80 \\
             \rowcolor[rgb]{0.882,0.882,0.882}\multirow{-2}{*}{FLAVA}\cellcolor[rgb]{1,1,1}&  \checkmark  & 57.20 & \underline{\textbf{54.09}} & \underline{\textbf{56.72}} & \underline{\textbf{55.23}} & \underline{\textbf{54.55}} & \underline{\textbf{56.14}} & \underline{\textbf{55.58}} & \underline{\textbf{54.07}} & \underline{\textbf{55.20}}\\ 
            
             &  & 
             \underline{56.76}&	50.85	&53.88&	52.05	&50.49&	53.89	&50.81	&49.87&	51.69
             \\
            \rowcolor[rgb]{0.882,0.882,0.882}\multirow{-2}{*}{\textbf{avg.}} \cellcolor[rgb]{1,1,1}&  \checkmark & 
            56.19&	\underline{51.22} & \underline{54.11} & \underline{52.35} & \underline{51.11} & \underline{53.93} & \underline{51.61} & \underline{50.60} & \underline{52.13}
            \\
             \toprule
            \end{tabular}
        
    \end{adjustbox}
    \caption{\textbf{Averaged translate-test results with different origins of training dataset}    
    Each accuracy represents the average of three training with different random seeds.
    The indicators are the same as Table~\ref{tab:translate-test_main}.
    }
    \label{tab:translate-test_three_times_run}
    \vspace{-3mm}
    \end{table*}

\begin{figure*}[t!]
\centering
\includegraphics[width=0.9\textwidth]
{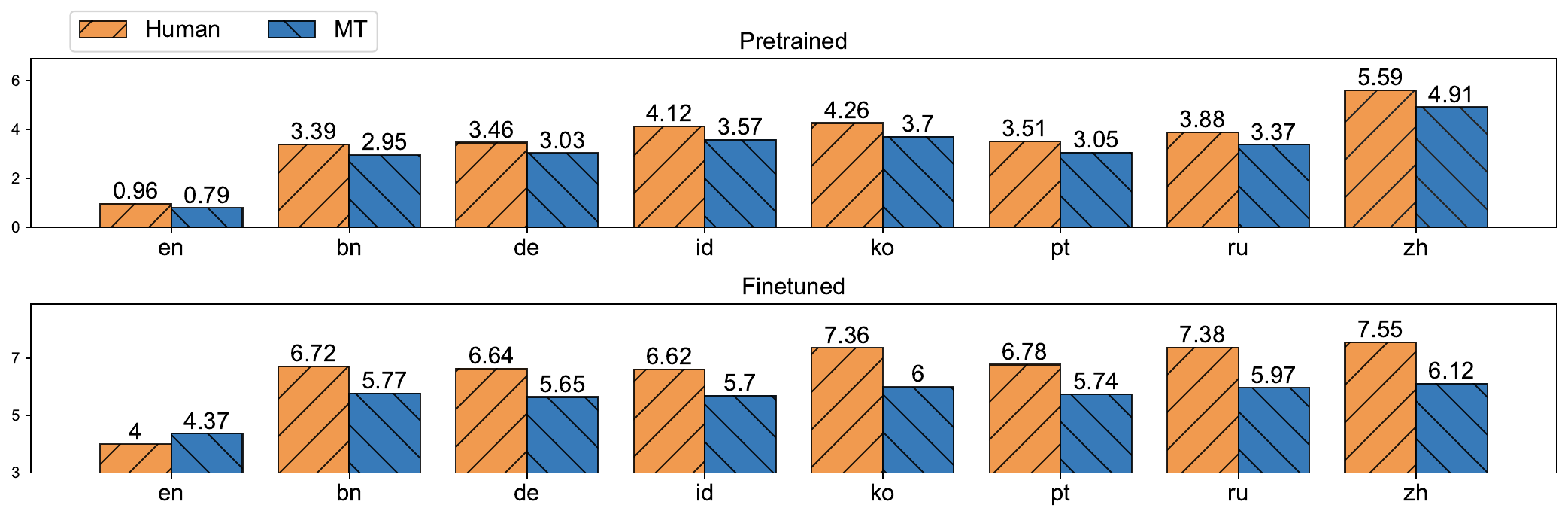}
\vspace{-3mm}
\caption{Representation discrepancy of translate-test evaluation samples against training samples from different data origins (Human and MT). Pretrained or finetuned VisualBERT is used to encode representation, and FID score is used as a distance metric. A lower score indicates a low distance between training and evaluation samples.}
\vspace{-3mm}
\label{fig:fid}
\end{figure*}
\noindent \textbf{Full results of Fig.~\ref{fig:fid_avg}}
Fig.~\ref{fig:fid} presents full results of Fig.~\ref{fig:fid_avg} across different target languages.

\noindent \textbf{Full results of Table~\ref{tab:varying_nmt}}
Table~\ref{tab:tab:varying_nmt_full} presents full results with varying MT systems for RT translation and a translate-test approach.

\begin{table*}[t]
\centering

\Rotatebox{90}{
\begin{adjustbox}{width=1.45\textwidth}
\begin{tabular}{|c|c|c|cccccccc|cccccccc|cccccccc|cccccccc|cccccccc|}
\hline
 & \textbf{} & en & bn & de & id & ko & pt & ru & zh & \textbf{avg.} &  bn & de & id & ko & pt & ru & zh & \textbf{avg.} & bn & de & id & ko & pt & ru & zh & \textbf{avg.} & bn & de & id & ko & pt & ru & zh & \textbf{avg.}&  bn & de & id & ko & pt & ru & zh & \textbf{avg.}\\ \hline
\textit{\textbf{Models}} & \textit{\textbf{}} & \textit{\textbf{}} & \multicolumn{8}{|c|}{\textbf{GMT}} & \multicolumn{8}{|c|}{\textbf{NLLB}} & \multicolumn{8}{|c|}{\textbf{NLLB\_3.3B}} & \multicolumn{8}{|c|}{\textbf{M2M-418M}} & \multicolumn{8}{|c|}{\textbf{M2M-1.2B}} \\
\hline
\multirow{5}{*}{MUNITER} & Human & 57.33 & 50.67 & 54.09 & 52.54 & 50.67 & 54.21 & 49.69 & 49.57 & 51.63 & 49.65 & 53.8 & 49.90 & 47.40 & 52.22 & 46.96 & 45.1 & 49.29 & 50.87 & 54.48 & 51.76 & 49.17 & 51.86 & 47.17 & 45.76 & 50.15 & 44.19 & 48.84 & 47.13 & 43.85 & 47.50 & 42.43 & 43.57 & 45.36 & 44.74 & 51.88 & 49.55 & 45.72 & 51.62 & 45.39 & 46.04 & 47.85 \\ 
 & NLLB-small & 55.38 & 51.85 & 55.15 & 53.36 & 52.7 & 54.7 & 52.77 & 52.27 & 53.26 & 51.27 & 54.82 & 51.97 & 50.17 & 54.43 & 52.65 & 48.59 & 51.99 & 52.23 & 54.67 & 53.09 & 50.99 & 54.12 & 53.21 & 49.44 & 52.54 & 46.09 & 51.71 & 49.56 & 47.37 & 51.44 & 49.36 & 48 & 49.08 & 46.84 & 53.36 & 51.41 & 48.47 & 52.91 & 51.22 & 49.24 & 50.49 \\
 & NLLB-large & 55.70 & 52.34 & 55.66 & 53.48 & 53.36 & 54.72 & 53.98 & 52.29 & 53.69 & 51.25 & 55.16 & 52.07 & 50.31 & 54.48 & 52.88 & 48.43 & 52.08 & 52.70 & 55.58 & 53.23 & 51.37 & 54.89 & 53.94 & 50.07 & 53.11 & 46.37 & 51.47 & 49.31 & 47.06 & 51.03 & 49.01 & 47.48 & 48.82 & 46.90 & 53.59 & 51.27 & 48.22 & 53.04 & 51.42 & 48.56 & 50.43 \\
 & M2M-small & 53.96 & 50.84 & 53.1 & 51.77 & 51.48 & 52.85 & 51.65 & 50.54 & 51.75 & 50.37 & 52.5 & 50.52 & 48.47 & 52.55 & 50.97 & 47.23 & 50.37 & 50.65 & 52.89 & 51.23 & 49.47 & 52.81 & 51.23 & 48.58 & 50.98 & 47.24 & 51.37 & 49.40 & 47.54 & 51.76 & 49.46 & 47.99 & 49.25 & 47.49 & 52.53 & 50.32 & 47.88 & 51.95 & 50.20 & 48.12 & 49.78 \\
 & M2M-large & \textbf{54.87} & \textbf{51.47} & \textbf{54.33} & \textbf{52.22} & \textbf{52.23} & \textbf{53.35} & \textbf{52.36} & \textbf{50.87} & \textbf{52.40} & \textbf{51.03} & 54.15 & 51.68 & 49.59 & 53.42 & 52.15 & 48.51 & 51.50 & 51.56 & 54.13 & 52.50 & 50.63 & 53.74 & 52.44 & 49.54 & 52.08 & 47.07 & 51.85 & 49.79 & 48.00 & 51.61 & 50.07 & 48.37 & 49.54 & 47.45 & 53.63 & 51.80 & 48.79 & 52.81 & 51.90 & 49.63 & 50.86 \\ \hline
 
\multirow{5}{*}{XUNITER} & Human & 56.98 & 50.76 & 54.63 & 52.37 & 50.52 & 54.24 & 48.91 & 49.94 & 51.62 & 49.60 & 53.59 & 49.57 & 47.11 & 51.24 & 45.33 & 45.29 & 48.82 & 50.69 & 54.09 & 51.36 & 48.86 & 51.20 & 45.77 & 46.06 & 49.72 & 43.82 & 48.79 & 46.80 & 43.16 & 44.58 & 37.05 & 43.66 & 43.98 & 44.08 & 51.61 & 48.76 & 45.44 & 50.52 & 41.49 & 46.02 & 46.85 \\
 & NLLB-small & 54.48 & 51.14 & 54.27 & 52.32 & 51.49 & 53.59 & 52.47 & 51.27 & 52.36 & 50.47 & 53.92 & 51.16 & 49.01 & 53.30 & 51.73 & 47.67 & 51.04 & 51.45 & 53.82 & 52.06 & 50.21 & 53.38 & 51.91 & 48.88 & 51.67 & 46.44 & 51.28 & 49.09 & 47.24 & 50.25 & 48.40 & 47.58 & 48.61 & 46.45 & 52.7 & 50.92 & 47.79 & 52.07 & 50.13 & 48.28 & 49.76 \\
 & NLLB-large & 55.22 & 52.10 & 54.97 & 52.66 & 52.51 & 54.18 & 52.85 & 52.23 & 53.07 & 51.00 & 54.64 & 51.42 & 49.78 & 53.94 & 52.00 & 48.22 & 51.57 & 51.95 & 54.74 & 52.62 & 51.01 & 54.25 & 52.90 & 49.68 & 52.45 & 46.33 & 51.11 & 49.10 & 47.69 & 50.37 & 48.29 & 47.71 & 48.66 & 46.70 & 53.26 & 51.03 & 48.43 & 52.68 & 50.30 & 48.87 & 50.18 \\
 & M2M-small & 53.40 & 50.93 & 52.67 & 50.87 & 50.6 & 52.33 & 51.33 & 50.12 & 51.26 & 49.97 & 52.27 & 50.12 & 47.65 & 51.92 & 50.22 & 47.42 & 49.94 & 50.49 & 52.23 & 50.64 & 49.02 & 52.02 & 50.51 & 48.12 & 50.43 & 47.02 & 51.07 & 48.95 & 46.82 & 50.88 & 49.11 & 47.73 & 48.80 & 46.95 & 51.9 & 49.92 & 47.54 & 51.34 & 49.55 & 47.79 & 49.28 \\
 & M2M-large & \textbf{54.42} & \textbf{52.05} & \textbf{54.26} & \textbf{52.31} & \textbf{51.86} & \textbf{53.71} & \textbf{53.19} & \textbf{50.75} & \textbf{52.59} & \textbf{50.68} & 53.95 & 51.32 & 48.71 & 52.95 & 51.85 & 47.8 & 51.04 & 51.03 & 53.74 & 52.00 & 49.71 & 53.32 & 51.98 & 48.94 & 51.53 & 47.31 & 51.92 & 50.10 & 48.26 & 51.56 & 50.26 & 48.51 & 49.70 & 46.97 & 53.45 & 51.19 & 48.87 & 52.92 & 51.76 & 49.32 & 50.64 \\ \hline
 
\multirow{5}{*}{UC$^2$} & Human & 56.85 & 51.34 & 54.01 & 52.35 & 50.75 & 53.81 & 51.93 & 50.04 & 52.03 & 49.73 & 53.08 & 49.75 & 47.51 & 52.77 & 49.90 & 45.76 & 49.79 & 51.42 & 54.13 & 51.64 & 49.32 & 53.35 & 51.08 & 47.48 & 51.20 & 44.17 & 49.17 & 47.31 & 44.78 & 49.69 & 47.17 & 45.59 & 46.84 & 44.44 & 51.37 & 49.17 & 46.16 & 51.63 & 49.25 & 47.06 & 48.44 \\
 & NLLB-small & 54.87 & 51.80 & 54.55 & 53.09 & 52.27 & 54.18 & 53.13 & 51.95 & 53.00 & 50.71 & 54.23 & 51.84 & 49.63 & 53.76 & 52.50 & 48.23 & 51.56 & 51.88 & 53.99 & 52.84 & 51.14 & 53.89 & 52.35 & 49.2 & 52.18 & 46.30 & 51.41 & 49.62 & 47.14 & 50.93 & 49.69 & 48.06 & 49.02 & 46.36 & 52.9 & 51.45 & 48.22 & 52.61 & 51.26 & 49.34 & 50.31 \\
 & NLLB-large & 55.12 & 52.35 & 55.1 & 53.29 & 53.07 & 54.17 & 53.36 & 52.73 & 53.44 & 51.07 & 54.44 & 51.77 & 50.04 & 53.98 & 52.75 & 48.58 & 51.80 & 52.42 & 54.7 & 53.10 & 51.14 & 54.05 & 53.59 & 49.71 & 52.67 & 46.14 & 51.38 & 49.40 & 47.86 & 50.48 & 49.32 & 47.87 & 48.92 & 47.07 & 53.2 & 51.11 & 48.82 & 52.82 & 51.13 & 49.28 & 50.49 \\
 & M2M-small & 53.50 & 50.86 & 53.04 & 51.32 & 50.99 & 52.61 & 51.78 & 50.88 & 51.64 & 50.18 & 52.73 & 50.50 & 48.67 & 52.19 & 50.89 & 48.37 & 50.50 & 50.35 & 52.5 & 51.02 & 49.43 & 52.12 & 50.77 & 49.13 & 50.76 & 47.51 & 51.3 & 49.58 & 47.26 & 51.19 & 49.44 & 48.01 & 49.18 & 47.45 & 51.94 & 50.41 & 47.83 & 51.77 & 50.75 & 48.67 & 49.83 \\
 & M2M-large & 54.21 & 52.07 & 53.76 & 52.23 & 52.29 & 53.25 & 52.62 & 50.62 & 52.41 & 50.79 & 53.29 & 51.45 & 49.68 & 52.71 & 51.87 & 48.31 & 51.16 & 51.07 & 53.24 & 51.78 & 50.23 & 52.62 & 52.08 & 49.26 & 51.47 & 47.10 & 51.3 & 50.52 & 48.44 & 51.43 & 50.56 & 48.39 & 49.68 & 47.35 & 52.81 & 51.34 & 48.82 & 52.69 & 51.74 & 49.6 & 50.62 \\ \hline

 \multirow{5}{*}{M$^3$P} & Human & 54.45 & 49.18 & 52.14 & 49.87 & 48.59 & 51.87 & 49.05 & 48.38 & 49.87 & 47.88 & 51.5 & 47.59 & 45.62 & 50.37 & 46.63 & 44.53 & 47.73 & 49.13 & 52.07 & 49.10 & 46.90 & 50.45 & 47.61 & 45.95 & 48.74 & 43.08 & 47.5 & 45.61 & 43.20 & 46.88 & 43.15 & 43.55 & 44.71 & 43.66 & 50.21 & 47.23 & 44.42 & 49.66 & 45.25 & 45.42 & 46.55 \\
 & NLLB-small & 51.88 & 49.65 & 52.07 & 50.32 & 49.25 & 51.19 & 50.63 & 49.48 & 50.37 & 48.65 & 51.77 & 48.87 & 47.08 & 50.96 & 49.73 & 46.07 & 49.02 & 49.44 & 51.3 & 49.64 & 48.23 & 51.05 & 49.86 & 46.8 & 49.47 & 44.51 & 49.63 & 47.25 & 44.73 & 48.53 & 47.19 & 45.48 & 46.76 & 44.91 & 50.85 & 49.05 & 45.59 & 50.11 & 48.90 & 46.61 & 48.00 \\
 & NLLB-large & 52.97 & 50.63 & 53.03 & 51.42 & 50.38 & 52.11 & 51.8 & 50.41 & 51.40 & 49.38 & 52.37 & 49.82 & 47.94 & 51.92 & 50.68 & 47.44 & 49.94 & 50.86 & 52.9 & 50.52 & 49.29 & 52.27 & 51.43 & 48.2 & 50.78 & 45.36 & 49.63 & 47.99 & 45.77 & 48.86 & 47.60 & 46.1 & 47.33 & 46.13 & 51.22 & 49.51 & 46.67 & 50.81 & 49.45 & 47.27 & 48.72 \\
 & M2M-small & 50.63 & 48.43 & 50.2 & 48.65 & 48.43 & 49.64 & 49.37 & 47.92 & 48.95 & 47.31 & 49.62 & 47.71 & 46.18 & 49.45 & 48.21 & 45.68 & 47.74 & 48.04 & 49.49 & 48.31 & 47.05 & 49.47 & 48.33 & 46.41 & 48.16 & 45.26 & 48.89 & 47.01 & 45.00 & 48.79 & 46.92 & 45.73 & 46.80 & 45.10 & 49.64 & 47.74 & 45.41 & 48.90 & 47.91 & 45.68 & 47.20 \\
 & M2M-large & 51.07 & 49.07 & 50.84 & 49.51 & 49.33 & 50.33 & 49.88 & 48.94 & 49.70 & 48.41 & 50.25 & 48.55 & 46.87 & 49.79 & 49.08 & 46.47 & 48.49 & 48.63 & 50.37 & 49.03 & 47.68 & 50.11 & 49.17 & 46.96 & 48.85 & 44.85 & 49.02 & 47.71 & 45.98 & 48.71 & 47.53 & 45.9 & 47.10 & 45.40 & 50.08 & 48.60 & 46.42 & 50.02 & 49.13 & 47.03 & 48.10 \\
 \hline

\multirow{5}{*}{LXMERT} & Human & 55.40 & 49.64 & 52.83 & 50.80 & 49.17 & 52.49 & 47.54 & 48.02 & 50.07 & 47.92 & 52.21 & 48.81 & 45.79 & 50.94 & 44.65 & 44.55 & 47.84 & 49.27 & 52.76 & 50.13 & 47.30 & 50.85 & 45.20 & 45.39 & 48.70 & 42.78 & 48.2 & 46.31 & 42.98 & 46.02 & 39.04 & 43.19 & 44.07 & 43.11 & 50.58 & 48.00 & 44.28 & 49.98 & 42.29 & 45.07 & 46.19 \\
 & NLLB-small & 52.95 & 50.00 & 52.85 & 50.91 & 50.14 & 52.08 & 50.97 & 50.11 & 51.01 & 49.32 & 52.86 & 49.81 & 47.59 & 51.95 & 50.46 & 46.74 & 49.82 & 50.38 & 52.58 & 50.70 & 49.01 & 51.88 & 50.54 & 47.36 & 50.35 & 44.55 & 50.06 & 47.56 & 45.71 & 49.07 & 47.35 & 46.14 & 47.21 & 45.46 & 51.53 & 49.07 & 46.47 & 50.75 & 48.39 & 47.04 & 48.39 \\
 & NLLB-large & 53.44 & 50.20 & 52.93 & 51.34 & 50.41 & 52.47 & 51.44 & 50.25 & 51.29 & 49.29 & 52.61 & 49.94 & 47.87 & 52.15 & 51.05 & 46.84 & 49.96 & 50.65 & 52.83 & 50.97 & 49.50 & 52.46 & 52.05 & 48.19 & 50.95 & 44.74 & 49.48 & 47.30 & 45.91 & 48.86 & 47.37 & 45.79 & 47.06 & 45.31 & 51.37 & 49.34 & 46.94 & 50.84 & 49.25 & 47.42 & 48.64 \\
 & M2M-small & 51.56 & 48.50 & 50.61 & 48.88 & 48.55 & 50.37 & 49.5 & 48.19 & 49.23 & 48.08 & 50.15 & 48.19 & 46.47 & 50.33 & 48.62 & 45.68 & 48.22 & 48.02 & 50.15 & 48.54 & 47.20 & 50.19 & 49.04 & 46.5 & 48.52 & 45.36 & 49.32 & 47.43 & 45.77 & 49.30 & 47.89 & 46.26 & 47.33 & 45.60 & 50.02 & 48.17 & 45.76 & 49.83 & 48.49 & 46.4 & 47.75 \\
 & M2M-large & 52.80 & 49.98 & 52.45 & 50.89 & 50.46 & 51.89 & 51.47 & 49.21 & 50.91 & 48.93 & 51.92 & 49.72 & 47.11 & 51.20 & 50.37 & 46.31 & 49.37 & 49.66 & 52.27 & 50.46 & 48.22 & 51.49 & 50.63 & 47.41 & 50.02 & 45.59 & 50.19 & 48.23 & 46.57 & 49.89 & 48.65 & 46.4 & 47.93 & 45.32 & 51.59 & 49.52 & 46.65 & 50.81 & 49.94 & 47.49 & 48.76 \\ \hline
 
\multirow{5}{*}{UNITER} & Human & 57.47 & 51.74 & 54.52 & 52.79 & 51.27 & 54.56 & 52.27 & 50.33 & 52.50 & 50.14 & 54.01 & 50.25 & 47.60 & 53.33 & 50.53 & 45.79 & 50.24 & 51.58 & 54.76 & 52.29 & 49.26 & 54.03 & 51.62 & 47.27 & 51.54 & 44.36 & 49.39 & 47.51 & 45.67 & 48.19 & 43.31 & 45.27 & 46.24 & 44.54 & 52.33 & 49.73 & 46.40 & 51.59 & 46.97 & 46.77 & 48.33 \\
 & NLLB-small & 55.49 & 52.03 & 54.76 & 53.28 & 52.66 & 54.44 & 53.18 & 52.02 & 53.20 & 51.54 & 54.98 & 51.93 & 49.64 & 54.53 & 53.03 & 48.86 & 52.07 & 52.45 & 54.9 & 52.87 & 51.26 & 54.38 & 53.36 & 49.52 & 52.68 & 46.73 & 51.98 & 49.99 & 47.92 & 51.29 & 49.21 & 48.36 & 49.35 & 47.17 & 53.58 & 51.52 & 48.48 & 53.29 & 51.36 & 49.6 & 50.71 \\
 & NLLB-large & 55.92 & 52.32 & 55.53 & 53.67 & 52.93 & 54.66 & 53.56 & 52.6 & 53.61 & 51.66 & 55.26 & 52.54 & 50.06 & 54.85 & 52.82 & 49.53 & 52.39 & 53.36 & 55.59 & 53.67 & 51.66 & 54.91 & 53.78 & 50.39 & 53.34 & 46.93 & 52.14 & 49.82 & 47.97 & 51.12 & 48.85 & 48 & 49.26 & 47.49 & 53.75 & 51.76 & 48.82 & 53.19 & 51.26 & 49.44 & 50.82 \\
 & M2M-small & 54.13 & 50.56 & 52.99 & 51.15 & 50.85 & 52.65 & 51.34 & 49.95 & 51.36 & 49.94 & 52.56 & 50.41 & 48.33 & 52.74 & 50.86 & 47.34 & 50.31 & 50.56 & 52.74 & 51.32 & 49.24 & 52.58 & 51.46 & 48.24 & 50.88 & 46.74 & 51.37 & 49.28 & 47.14 & 51.22 & 49.75 & 47.46 & 48.99 & 46.61 & 52.15 & 50.38 & 47.75 & 51.98 & 50.45 & 48.07 & 49.63 \\
 & M2M-large & 55.35 & 52.27 & 54.64 & 52.97 & 52.79 & 53.88 & 53.24 & 51.44 & 53.03 & 51.25 & 54.09 & 51.96 & 49.55 & 53.66 & 52.39 & 48.67 & 51.65 & 52.26 & 54.46 & 52.74 & 50.80 & 53.83 & 53.00 & 49.52 & 52.37 & 47.42 & 52.46 & 50.68 & 48.84 & 51.92 & 50.78 & 48.86 & 50.14 & 47.87 & 53.97 & 51.97 & 49.52 & 53.51 & 52.33 & 49.85 & 51.29 \\ \hline
 
\multirow{5}{*}{VILBERT} & Human & 56.72 & 50.84 & 54.1 & 52.27 & 50.73 & 53.98 & 49.91 & 49.92 & 51.68 & 49.24 & 53.64 & 49.83 & 47.19 & 51.88 & 47.11 & 45.2 & 49.16 & 50.64 & 54.09 & 51.53 & 48.86 & 52.42 & 48.21 & 46.3 & 50.29 & 44.08 & 49.41 & 47.00 & 43.97 & 46.84 & 39.92 & 44.43 & 45.09 & 44.46 & 51.76 & 48.88 & 45.68 & 50.68 & 43.38 & 46.16 & 47.29 \\
 & NLLB-small & 54.86 & 51.40 & 54.32 & 52.73 & 51.88 & 54.21 & 52.57 & 51.66 & 52.68 & 50.95 & 54.22 & 51.73 & 49.24 & 53.82 & 52.08 & 48.04 & 51.44 & 51.91 & 54.33 & 52.39 & 51.28 & 53.85 & 52.23 & 48.86 & 52.12 & 46.76 & 51.6 & 49.24 & 47.30 & 50.56 & 48.74 & 48.01 & 48.89 & 47.17 & 52.85 & 51.20 & 47.86 & 52.40 & 50.51 & 48.77 & 50.11 \\
 & NLLB-large & 55.22 & 52.23 & 54.85 & 53.43 & 52.75 & 54.26 & 53.69 & 52.22 & 53.35 & 51.22 & 54.48 & 52.21 & 49.82 & 54.01 & 52.69 & 48.96 & 51.91 & 52.58 & 54.8 & 52.84 & 51.40 & 54.13 & 53.36 & 49.73 & 52.69 & 46.37 & 51.08 & 49.53 & 47.18 & 50.62 & 48.67 & 47.96 & 48.77 & 47.07 & 53.12 & 51.27 & 48.26 & 52.48 & 51.49 & 49.27 & 50.42 \\
 & M2M-small & 53.52 & 51.07 & 52.9 & 51.37 & 50.97 & 52.36 & 51.57 & 50.43 & 51.52 & 50.41 & 52.27 & 50.53 & 48.52 & 52.44 & 50.56 & 47.61 & 50.33 & 50.77 & 52.81 & 51.35 & 49.70 & 52.28 & 51.26 & 48.69 & 50.98 & 47.35 & 50.94 & 49.63 & 47.11 & 50.87 & 49.37 & 47.97 & 49.03 & 47.34 & 52.08 & 50.79 & 47.35 & 51.50 & 50.33 & 48.31 & 49.67 \\
 & M2M-large & 55.12 & 52.15 & 54.32 & 52.66 & 52.21 & 53.6 & 53.43 & 51.44 & 52.83 & 51.11 & 53.66 & 51.72 & 49.56 & 52.81 & 52.25 & 48.12 & 51.32 & 51.25 & 53.94 & 52.10 & 50.45 & 53.20 & 52.27 & 49.12 & 51.76 & 47.27 & 52.39 & 49.97 & 48.33 & 51.53 & 50.55 & 48.71 & 49.82 & 47.48 & 53.44 & 51.83 & 48.74 & 52.67 & 52.22 & 49.46 & 50.83 \\  \hline
 
\multirow{5}{*}{VisualBERT} & Human & 55.17 & 49.43 & 52.58 & 50.34 & 48.66 & 52.72 & 50.5 & 48.89 & 50.45 & 48.08 & 51.92 & 48.19 & 45.84 & 51.87 & 48.96 & 44.56 & 48.49 & 49.64 & 52.42 & 49.65 & 47.51 & 52.51 & 49.75 & 45.48 & 49.57 & 43.08 & 47.92 & 45.36 & 43.13 & 45.67 & 40.62 & 43.43 & 44.17 & 43.34 & 50.29 & 47.21 & 44.39 & 49.70 & 44.47 & 44.86 & 46.32 \\
 & NLLB-small & 53.59 & 50.60 & 52.87 & 51.00 & 50.68 & 52.77 & 51.11 & 50.52 & 51.36 & 48.99 & 52.68 & 49.69 & 47.30 & 52.35 & 50.65 & 46.72 & 49.77 & 50.47 & 52.89 & 50.53 & 49.11 & 52.81 & 51.30 & 47.58 & 50.67 & 44.54 & 50.09 & 47.42 & 45.55 & 49.16 & 47.25 & 46.18 & 47.17 & 45.50 & 51.32 & 48.77 & 46.33 & 51.08 & 48.82 & 47.27 & 48.44 \\
 & NLLB-large & 53.51 & 50.57 & 53.1 & 51.17 & 50.45 & 52.59 & 51.47 & 50.97 & 51.47 & 49.44 & 52.73 & 49.52 & 47.87 & 52.55 & 50.87 & 46.94 & 49.99 & 50.92 & 53.29 & 50.90 & 49.10 & 52.42 & 51.60 & 48.25 & 50.93 & 45.08 & 49.3 & 47.41 & 45.92 & 48.47 & 46.70 & 45.95 & 46.98 & 45.17 & 51.22 & 49.04 & 46.58 & 50.56 & 49.04 & 47.53 & 48.45 \\
 & M2M-small & 52.26 & 49.36 & 51.47 & 49.81 & 49.35 & 51.19 & 50.31 & 49.35 & 50.12 & 48.61 & 51.05 & 48.76 & 47.42 & 50.86 & 48.94 & 46.42 & 48.87 & 49.16 & 51.15 & 49.26 & 48.00 & 50.72 & 49.59 & 47.27 & 49.31 & 45.46 & 49.86 & 47.94 & 46.03 & 49.84 & 48.22 & 46.64 & 47.71 & 45.79 & 50.29 & 48.59 & 45.88 & 50.45 & 48.66 & 46.62 & 48.04 \\
 & M2M-large & 52.71 & 50.10 & 52.56 & 50.35 & 49.9 & 52.01 & 51.25 & 49.95 & 50.87 & 48.83 & 52.15 & 49.44 & 47.61 & 51.52 & 50.24 & 47.05 & 49.55 & 50.02 & 52.42 & 49.98 & 48.63 & 51.60 & 50.62 & 47.8 & 50.15 & 46.02 & 50.26 & 48.09 & 46.60 & 49.76 & 48.07 & 46.85 & 47.95 & 46.00 & 51.73 & 49.48 & 46.98 & 51.32 & 49.86 & 48.12 & 49.07 \\ \hline
 
\multirow{5}{*}{VL-BERT} & Human & 57.79 & 51.22 & 54.47 & 52.62 & 50.94 & 54.79 & 51.17 & 50.02 & 52.18 & 50.14 & 53.67 & 50.60 & 47.82 & 53.55 & 48.78 & 45.4 & 49.99 & 51.38 & 54.78 & 52.22 & 49.61 & 53.51 & 49.37 & 46.68 & 51.08 & 44.20 & 48.8 & 46.92 & 44.08 & 47.59 & 41.04 & 44.01 & 45.23 & 44.35 & 51.89 & 49.71 & 46.31 & 51.69 & 45.85 & 46.57 & 48.05 \\
 & NLLB-small & 55.91 & 52.41 & 55.76 & 53.84 & 52.9 & 55.26 & 53.56 & 52.5 & 53.75 & 51.79 & 55.47 & 52.19 & 50.12 & 54.90 & 52.98 & 48.81 & 52.32 & 52.91 & 55.4 & 53.35 & 51.68 & 55.07 & 53.40 & 49.83 & 53.09 & 47.15 & 52.08 & 50.01 & 47.80 & 51.26 & 49.39 & 48.37 & 49.44 & 47.78 & 53.78 & 51.99 & 48.54 & 53.08 & 50.91 & 49.15 & 50.75 \\
 & NLLB-large & 55.61 & 52.38 & 55.27 & 53.43 & 52.58 & 54.63 & 53.32 & 52.31 & 53.42 & 51.20 & 54.78 & 51.92 & 49.79 & 54.55 & 52.92 & 48.78 & 51.99 & 52.84 & 55.06 & 53.13 & 51.47 & 54.55 & 53.78 & 49.58 & 52.92 & 46.63 & 51.04 & 49.55 & 47.64 & 50.65 & 48.58 & 47.84 & 48.85 & 47.06 & 53.04 & 51.07 & 48.67 & 52.67 & 51.16 & 49.4 & 50.44 \\
 & M2M-small & 54.67 & 51.43 & 53.79 & 52.04 & 51.45 & 53.3 & 52.05 & 50.92 & 52.14 & 50.48 & 53.09 & 51.14 & 49.13 & 53.03 & 51.25 & 48.08 & 50.89 & 51.16 & 53.28 & 51.38 & 49.91 & 53.10 & 52.03 & 49.06 & 51.42 & 47.92 & 51.5 & 49.77 & 47.77 & 52.04 & 50.17 & 48.31 & 49.64 & 48.23 & 52.53 & 51.26 & 48.59 & 52.58 & 51.14 & 48.64 & 50.42 \\
 & M2M-large & 55.64 & 52.23 & 54.71 & 53.09 & 52.3 & 54.64 & 53.2 & 51.14 & 53.04 & 51.17 & 54.61 & 52.22 & 49.55 & 53.75 & 52.47 & 48.31 & 51.73 & 51.64 & 54.78 & 52.46 & 50.59 & 53.74 & 52.89 & 49.37 & 52.21 & 47.35 & 52.46 & 50.59 & 48.31 & 52.02 & 50.72 & 48.62 & 50.01 & 47.31 & 54.28 & 51.80 & 49.25 & 53.30 & 52.50 & 49.3 & 51.11 \\ \hline

\multirow{5}{*}{BLIP-2} & Human & 58.05 & 52.03 & 54.7 & 52.99 & 51.57 & 54.91 & 52.36 & 51.22 & 53.48 & 50.13 & 54.4 & 50.82 & 47.94 & 53.94 & 50.43 & 46.78 & 50.63 & 52.32 & 55.14 & 52.39 & 49.75 & 54.68 & 51.12 & 47.98 & 51.91 & 44.81 & 49.45 & 47.61 & 45.10 & 48.96 & 46.27 & 45.81 & 46.86 & 45.36 & 52.64 & 49.54 & 46.61 & 51.98 & 48.70 & 47.62 & 48.92 \\
 & NLLB-small & 55.69 & 52.86 & 55.15 & 53.58 & 52.66 & 54.69 & 53.93 & 52.85 & 53.93 & 52.18 & 55.41 & 52.25 & 49.83 & 54.79 & 53.59 & 49.64 & 52.53 & 53.16 & 55.29 & 53.16 & 51.38 & 54.82 & 53.57 & 50 & 53.05 & 47.18 & 52.11 & 50.32 & 47.85 & 51.62 & 50.12 & 48.38 & 49.65 & 47.29 & 53.94 & 51.63 & 48.76 & 53.26 & 51.46 & 49.78 & 50.87 \\
 & NLLB-large & 56.11 & 53.18 & 55.7 & 53.98 & 53.51 & 55.11 & 54.25 & 53.31 & 54.39 & 52.21 & 55.35 & 52.73 & 50.29 & 54.96 & 53.24 & 49.21 & 52.57 & 53.74 & 56.07 & 53.70 & 51.96 & 55.15 & 54.37 & 50.8 & 53.68 & 46.98 & 52.03 & 49.97 & 48.28 & 51.52 & 49.40 & 48.52 & 49.53 & 47.81 & 54.08 & 51.62 & 48.76 & 53.32 & 51.66 & 49.9 & 51.02 \\
 & M2M-small & 53.97 & 50.29 & 52.7 & 50.97 & 50.37 & 52.54 & 51.15 & 50.31 & 51.54 & 50.09 & 52.77 & 50.33 & 48.72 & 52.57 & 51.10 & 48.08 & 50.52 & 50.76 & 53.07 & 51.08 & 49.48 & 52.85 & 51.49 & 49.17 & 51.13 & 47.23 & 51.39 & 49.60 & 47.23 & 51.42 & 49.59 & 48.5 & 49.28 & 47.30 & 52.24 & 50.67 & 47.62 & 51.84 & 50.50 & 48.47 & 49.81 \\
 & M2M-large & 55.41 & 52.03 & 54.52 & 52.77 & 52.71 & 54.04 & 53.32 & 51.95 & 53.34 & 51.07 & 54.41 & 51.80 & 49.86 & 54.19 & 52.99 & 49.34 & 51.95 & 52.27 & 54.44 & 52.79 & 51.25 & 54.07 & 53.00 & 50.41 & 52.60 & 47.68 & 52.77 & 50.49 & 48.47 & 51.76 & 51.31 & 49.52 & 50.29 & 48.27 & 53.99 & 52.05 & 49.46 & 53.10 & 52.40 & 50.58 & 51.41 \\ \hline
 
\multirow{5}{*}{InstructBLIP} & Human & 57.85 & 51.80 & 54.91 & 53.01 & 51.29 & 54.85 & 53.16 & 51.34 & 52.91 & 50.23 & 54.27 & 50.48 & 48.04 & 54.33 & 51.49 & 46.63 & 50.78 & 51.81 & 54.77 & 52.08 & 49.44 & 54.76 & 52.27 & 48.23 & 51.91 & 44.48 & 49.89 & 47.65 & 44.91 & 49.72 & 48.33 & 46.3 & 47.33 & 44.43 & 52.73 & 49.61 & 46.09 & 52.24 & 49.71 & 47.81 & 48.95 \\
 & NLLB-small & 55.61 & 52.27 & 54.71 & 53.57 & 52.55 & 54.57 & 53.67 & 52.52 & 53.41 & 51.77 & 55.14 & 52.27 & 49.89 & 54.56 & 53.34 & 49.29 & 52.32 & 52.76 & 54.99 & 53.27 & 51.50 & 54.79 & 53.65 & 50.29 & 53.04 & 46.76 & 52.2 & 50.50 & 47.42 & 51.26 & 50.49 & 48.34 & 49.57 & 47.30 & 53.59 & 51.49 & 48.18 & 53.04 & 51.43 & 49.63 & 50.67 \\
 & NLLB-large & 55.84 & 53.04 & 55.06 & 53.82 & 53.17 & 54.32 & 54.08 & 53.18 & 53.81 & 51.66 & 54.87 & 52.35 & 50.71 & 54.40 & 53.43 & 49.68 & 52.44 & 53.50 & 55.42 & 53.51 & 51.93 & 54.48 & 53.78 & 50.74 & 53.34 & 47.08 & 51.53 & 49.98 & 48.44 & 51.06 & 49.29 & 48.83 & 49.46 & 47.27 & 53.42 & 51.46 & 48.78 & 52.73 & 51.50 & 49.75 & 50.70 \\
 & M2M-small & 54.13 & 50.70 & 53.03 & 51.56 & 50.99 & 52.44 & 51.57 & 50.97 & 51.61 & 50.14 & 52.73 & 51.14 & 49.52 & 52.41 & 50.94 & 48.39 & 50.75 & 51.01 & 52.97 & 51.34 & 49.82 & 52.69 & 51.29 & 49.55 & 51.24 & 47.59 & 51.36 & 50.21 & 47.67 & 51.88 & 49.95 & 48.7 & 49.62 & 47.84 & 52.34 & 51.07 & 47.57 & 52.15 & 50.76 & 48.99 & 50.10 \\
 & M2M-large & 54.86 & 52.00 & 54.21 & 52.81 & 52.33 & 53.78 & 53.36 & 51.82 & 52.90 & 51.15 & 53.84 & 52.05 & 49.87 & 53.49 & 52.66 & 48.86 & 51.70 & 52.14 & 54.05 & 52.42 & 50.83 & 53.70 & 52.78 & 49.75 & 52.24 & 47.27 & 52.34 & 50.69 & 48.70 & 51.73 & 50.87 & 48.76 & 50.05 & 47.77 & 53.7 & 51.70 & 49.26 & 52.93 & 52.68 & 49.81 & 51.12 \\ \hline
 
\multirow{5}{*}{FLAVA} & Human & 58.84 & 53.47 & 56.26 & 54.11 & 52.85 & 55.84 & 53.64 & 52.18 & 54.65 & 51.74 & 54.93 & 51.69 & 49.13 & 54.91 & 51.26 & 47.23 & 51.56 & 52.94 & 55.76 & 53.13 & 51.03 & 55.13 & 52.17 & 48.53 & 52.67 & 45.37 & 50.16 & 48.86 & 46.87 & 49.90 & 46.76 & 46.48 & 47.77 & 46.14 & 52.99 & 50.72 & 47.81 & 52.74 & 49.17 & 48.4 & 49.71 \\
 & NLLB-small & 56.50 & 53.64 & 55.84 & 54.42 & 53.66 & 55.17 & 54.67 & 53.18 & 54.64 & 52.31 & 55.71 & 53.08 & 50.82 & 55.06 & 53.43 & 49.05 & 52.78 & 53.39 & 55.76 & 53.92 & 52.14 & 55.36 & 54.25 & 50.38 & 53.60 & 46.80 & 52.8 & 50.82 & 48.99 & 51.72 & 50.14 & 48.84 & 50.02 & 47.77 & 54.24 & 52.57 & 49.59 & 53.97 & 52.08 & 50.18 & 51.49 \\
 & NLLB-large & 56.87 & 53.94 & 56.35 & 54.99 & 54.51 & 55.96 & 55.61 & 53.82 & 55.26 & 52.55 & 55.95 & 52.86 & 50.92 & 55.45 & 54.19 & 49.82 & 53.11 & 53.50 & 55.42 & 53.51 & 51.93 & 54.48 & 53.78 & 50.74 & 53.34 & 47.08 & 51.53 & 49.98 & 48.44 & 51.06 & 49.29 & 48.83 & 49.46 & 47.53 & 52.27 & 50.75 & 49.13 & 52.19 & 50.41 & 49.13 & 50.20 \\
 & M2M-small & 55.80 & 53.16 & 54.83 & 53.66 & 53.11 & 54.51 & 53.59 & 52.66 & 53.92 & 51.94 & 54.43 & 52.68 & 50.37 & 54.11 & 52.66 & 49.65 & 52.26 & 52.39 & 54.55 & 53.22 & 51.33 & 54.44 & 52.85 & 50.37 & 52.74 & 48.57 & 52.96 & 50.83 & 48.95 & 52.53 & 50.63 & 49.17 & 50.52 & 48.74 & 53.45 & 52.13 & 49.26 & 53.20 & 51.90 & 49.63 & 51.19 \\
 & M2M-large & 55.35 & 52.97 & 54.68 & 53.39 & 53.37 & 54.29 & 53.96 & 51.7 & 53.71 & 51.79 & 54 & 52.13 & 50.21 & 53.47 & 52.56 & 48.86 & 51.86 & 52.05 & 54.31 & 52.76 & 51.51 & 53.86 & 53.15 & 50.06 & 52.53 & 47.64 & 52.5 & 50.78 & 49.20 & 51.84 & 50.88 & 48.7 & 50.22 & 47.75 & 53.77 & 52.12 & 49.51 & 53.12 & 52.70 & 50.26 & 51.32 \\ \hline
 
\end{tabular}
\end{adjustbox}
}

\caption{Full results of translate-test evaluation with different MT systems for RT-translation and translating evaluation samples in target languages into English. The averaged results are in Table~\ref{tab:varying_nmt}.}
\label{tab:tab:varying_nmt_full}
\end{table*}

\noindent \textbf{Full results of Table~\ref{tab:varied_pivot}}
Table~\ref{tab:pivot_full} presents full results with varying pivot languages used in RT translation. \textsc{NLLB-200-3.3B} is used as an MT system.

\begin{table*}[t!]
\small
\centering

\begin{adjustbox}{width=\textwidth}
\begin{tabular}{|c|c|c|cccccccc|cccccccc|}
\hline
\multirow{2}{*}{\textbf{Models}} & \multirow{2}{*}{\centering \begin{tabular}[c]{@{}l@{}}RT\\Pivot\end{tabular}} &  \multirow{2}{*}{en} & bn & de & id & ko & pt & ru & zh & \textbf{avg.} & bn & de & id & ko & pt & ru & zh & \textbf{avg.} \\ \cline{4-19}
 & & & \multicolumn{8}{c}{\textbf{GMT}} & \multicolumn{8}{|c|}{\textbf{NLLB-3.3B}} \\ \hline
  &
  bn &
  54.08 &
  51.89 &
  53.96 &
  52.46 &
  52.35 &
  53.01 &
  52.60 &
  52.30 &
  52.65 &
  52.54 &
  53.84 &
  52.85 &
  51.20 &
  52.97 &
  52.85 &
  50.30 &
  52.36 \\
 &
  de &
  55.70 &
  52.34 &
  55.66 &
  53.48 &
  53.36 &
  54.72 &
  53.98 &
  52.29 &
  53.69 &
  52.70 &
  55.58 &
  53.23 &
  51.37 &
  54.89 &
  53.94 &
  50.07 &
  53.11 \\
 &
  id &
  55.13 &
  52.24 &
  54.44 &
  54.02 &
  53.47 &
  54.31 &
  53.93 &
  52.94 &
  53.62 &
  53.08 &
  54.55 &
  54.66 &
  51.85 &
  54.35 &
  53.84 &
  51.14 &
  53.35 \\
 &
  ko &
  53.90 &
  51.30 &
  53.27 &
  52.31 &
  52.81 &
  53.07 &
  52.79 &
  52.11 &
  52.52 &
  51.65 &
  53.19 &
  52.54 &
  52.12 &
  53.24 &
  52.59 &
  50.68 &
  52.29 \\
 &
  pt &
  56.23 &
  52.43 &
  55.49 &
  53.67 &
  52.72 &
  55.72 &
  54.00 &
  52.97 &
  53.86 &
  53.08 &
  55.52 &
  53.74 &
  51.59 &
  55.97 &
  54.44 &
  50.39 &
  53.53 \\
 &
  ru &
  55.29 &
  51.53 &
  54.33 &
  52.82 &
  53.21 &
  53.88 &
  53.52 &
  52.56 &
  53.12 &
  52.39 &
  54.48 &
  53.47 &
  51.59 &
  54.23 &
  54.60 &
  50.52 &
  53.04 \\
\multirow{-7}{*}{MUNITER} &
  zh &
  53.06 &
  50.99 &
  52.77 &
  51.96 &
  52.07 &
  52.30 &
  51.92 &
  52.05 &
  52.01 &
  49.53 &
  50.75 &
  49.88 &
  48.88 &
  50.49 &
  49.67 &
  49.58 &
  49.83 \\ \hline
 &
  bn &
  54.23 &
  52.39 &
  53.68 &
  52.70 &
  52.22 &
  53.45 &
  53.18 &
  52.07 &
  52.81 &
  52.66 &
  53.30 &
  52.52 &
  51.14 &
  53.01 &
  52.43 &
  50.42 &
  52.21 \\
 &
  de &
  55.22 &
  52.10 &
  54.97 &
  52.66 &
  52.51 &
  54.18 &
  52.85 &
  52.23 &
  53.07 &
  51.95 &
  54.74 &
  52.62 &
  51.01 &
  54.25 &
  52.90 &
  49.68 &
  52.45 \\
 &
  id &
  54.83 &
  52.47 &
  54.05 &
  54.09 &
  53.21 &
  54.06 &
  53.42 &
  53.24 &
  53.51 &
  52.82 &
  53.54 &
  53.99 &
  51.30 &
  53.84 &
  52.89 &
  50.62 &
  52.71 \\
 &
  ko &
  53.55 &
  51.35 &
  52.79 &
  51.74 &
  52.15 &
  52.63 &
  52.18 &
  51.70 &
  52.08 &
  51.38 &
  52.69 &
  51.77 &
  51.34 &
  52.74 &
  51.65 &
  49.98 &
  51.65 \\
 &
  pt &
  55.51 &
  51.82 &
  54.55 &
  53.05 &
  52.15 &
  54.88 &
  53.74 &
  52.37 &
  53.22 &
  52.29 &
  54.45 &
  52.50 &
  50.87 &
  54.78 &
  53.47 &
  48.85 &
  52.46 \\
 &
  ru &
  54.77 &
  51.96 &
  54.05 &
  53.25 &
  52.70 &
  53.81 &
  53.72 &
  52.50 &
  53.14 &
  52.00 &
  53.91 &
  53.02 &
  50.91 &
  53.86 &
  53.92 &
  50.11 &
  52.53 \\
\multirow{-7}{*}{XUNITER} &
  zh &
  52.31 &
  50.99 &
  52.32 &
  51.22 &
  51.16 &
  51.84 &
  51.50 &
  51.68 &
  51.53 &
  48.20 &
  48.81 &
  48.23 &
  47.79 &
  48.88 &
  48.24 &
  47.96 &
  48.30 \\ \hline
 &
  bn &
  53.95 &
  52.51 &
  53.51 &
  52.73 &
  52.41 &
  53.19 &
  52.78 &
  52.20 &
  52.76 &
  52.62 &
  53.24 &
  52.42 &
  51.27 &
  52.91 &
  52.63 &
  50.73 &
  52.26 \\
 &
  de &
  55.12 &
  52.35 &
  55.10 &
  53.29 &
  53.07 &
  54.17 &
  53.36 &
  52.73 &
  53.44 &
  52.42 &
  54.70 &
  53.10 &
  51.14 &
  54.05 &
  53.59 &
  49.71 &
  52.67 \\
 &
  id &
  54.89 &
  53.12 &
  54.31 &
  54.29 &
  53.69 &
  54.14 &
  53.90 &
  53.38 &
  53.83 &
  52.77 &
  53.76 &
  53.77 &
  51.89 &
  53.85 &
  53.33 &
  51.04 &
  52.92 \\
 &
  ko &
  53.54 &
  51.69 &
  53.01 &
  52.50 &
  52.71 &
  52.88 &
  52.34 &
  52.02 &
  52.45 &
  51.83 &
  52.62 &
  52.13 &
  51.82 &
  52.62 &
  52.01 &
  50.33 &
  51.91 \\
 &
  pt &
  55.31 &
  52.16 &
  54.64 &
  52.89 &
  52.57 &
  54.65 &
  53.49 &
  52.02 &
  53.20 &
  52.33 &
  54.27 &
  53.01 &
  50.83 &
  54.85 &
  53.57 &
  49.00 &
  52.55 \\
 &
  ru &
  55.17 &
  52.67 &
  54.79 &
  53.22 &
  53.47 &
  54.31 &
  54.48 &
  52.70 &
  53.66 &
  52.71 &
  54.33 &
  53.55 &
  51.57 &
  54.09 &
  54.42 &
  51.07 &
  53.11 \\
\multirow{-7}{*}{UC$^2$} &
  zh &
  52.72 &
  51.04 &
  52.86 &
  51.84 &
  52.00 &
  52.39 &
  52.18 &
  51.88 &
  52.03 &
  48.62 &
  49.43 &
  49.07 &
  48.24 &
  49.48 &
  48.79 &
  48.80 &
  48.92 \\ \hline
 &
  bn &
  51.64 &
  50.13 &
  51.03 &
  49.83 &
  49.67 &
  50.47 &
  50.33 &
  49.73 &
  50.17 &
  50.25 &
  50.78 &
  49.56 &
  48.67 &
  50.21 &
  50.14 &
  48.20 &
  49.69 \\
 &
  de &
  52.97 &
  50.63 &
  53.03 &
  51.42 &
  50.38 &
  52.11 &
  51.80 &
  50.41 &
  51.40 &
  50.86 &
  52.90 &
  50.52 &
  49.29 &
  52.27 &
  51.43 &
  48.20 &
  50.78 \\
 &
  id &
  51.03 &
  49.10 &
  50.00 &
  50.29 &
  49.79 &
  50.40 &
  49.79 &
  49.55 &
  49.85 &
  49.16 &
  49.67 &
  49.86 &
  48.06 &
  50.19 &
  49.66 &
  47.54 &
  49.16 \\
 &
  ko &
  51.30 &
  49.23 &
  50.70 &
  49.56 &
  50.14 &
  50.35 &
  49.70 &
  49.79 &
  49.92 &
  49.15 &
  50.25 &
  49.22 &
  49.27 &
  50.06 &
  49.71 &
  48.35 &
  49.43 \\
 &
  pt &
  53.69 &
  50.60 &
  52.95 &
  51.34 &
  50.44 &
  52.98 &
  51.49 &
  50.55 &
  51.48 &
  50.72 &
  52.47 &
  50.59 &
  48.79 &
  52.73 &
  51.19 &
  48.21 &
  50.67 \\
 &
  ru &
  52.72 &
  50.14 &
  52.26 &
  50.45 &
  50.14 &
  51.61 &
  51.65 &
  50.38 &
  50.95 &
  50.81 &
  52.15 &
  50.52 &
  48.89 &
  51.57 &
  51.84 &
  48.44 &
  50.60 \\
\multirow{-7}{*}{M$^3$P} &
  zh &
  49.70 &
  48.16 &
  49.23 &
  48.78 &
  48.49 &
  48.86 &
  48.68 &
  48.81 &
  48.72 &
  48.00 &
  49.23 &
  48.33 &
  47.64 &
  48.56 &
  48.46 &
  47.92 &
  48.31 \\ \hline
 &
  bn &
  51.94 &
  50.42 &
  51.33 &
  50.47 &
  50.36 &
  51.16 &
  50.58 &
  50.10 &
  50.63 &
  50.72 &
  51.20 &
  50.76 &
  49.33 &
  51.09 &
  50.99 &
  48.76 &
  50.41 \\
 &
  de &
  53.44 &
  50.20 &
  52.93 &
  51.34 &
  50.41 &
  52.47 &
  51.44 &
  50.25 &
  51.29 &
  50.65 &
  52.83 &
  50.97 &
  49.50 &
  52.46 &
  52.05 &
  48.19 &
  50.95 \\
 &
  id &
  53.01 &
  50.44 &
  51.76 &
  52.16 &
  51.29 &
  52.11 &
  51.46 &
  50.99 &
  51.46 &
  50.61 &
  51.78 &
  51.77 &
  49.86 &
  52.19 &
  51.47 &
  49.47 &
  51.02 \\
 &
  ko &
  51.84 &
  49.61 &
  51.08 &
  50.63 &
  50.72 &
  51.00 &
  50.39 &
  50.20 &
  50.52 &
  50.09 &
  51.01 &
  50.63 &
  50.33 &
  50.87 &
  50.01 &
  48.77 &
  50.24 \\
 &
  pt &
  53.82 &
  50.42 &
  53.02 &
  51.64 &
  50.53 &
  53.07 &
  51.69 &
  50.56 &
  51.56 &
  50.65 &
  52.76 &
  51.01 &
  49.43 &
  53.32 &
  51.61 &
  48.02 &
  50.97 \\
 &
  ru &
  53.03 &
  49.72 &
  52.08 &
  50.99 &
  50.75 &
  51.84 &
  51.22 &
  50.42 &
  51.00 &
  50.54 &
  51.99 &
  51.11 &
  49.75 &
  51.91 &
  52.14 &
  48.79 &
  50.89 \\
\multirow{-7}{*}{LXMERT} &
  zh &
  51.02 &
  49.09 &
  50.09 &
  49.90 &
  49.76 &
  50.12 &
  49.85 &
  49.93 &
  49.82 &
  46.67 &
  47.45 &
  46.98 &
  46.57 &
  47.54 &
  46.75 &
  46.84 &
  46.97 \\ \hline
 &
  bn &
  54.15 &
  52.09 &
  53.26 &
  52.75 &
  52.07 &
  52.66 &
  52.68 &
  51.96 &
  52.50 &
  52.62 &
  52.90 &
  52.89 &
  50.98 &
  52.99 &
  52.74 &
  50.65 &
  52.25 \\
 &
  de &
  55.92 &
  52.32 &
  55.53 &
  53.67 &
  52.93 &
  54.66 &
  53.56 &
  52.60 &
  53.61 &
  53.36 &
  55.59 &
  53.67 &
  51.66 &
  54.91 &
  53.78 &
  50.39 &
  53.34 \\
 &
  id &
  55.61 &
  52.66 &
  54.33 &
  54.78 &
  54.00 &
  54.38 &
  53.76 &
  53.47 &
  53.91 &
  52.92 &
  54.42 &
  54.33 &
  52.16 &
  54.26 &
  54.23 &
  50.74 &
  53.29 \\
 &
  ko &
  53.90 &
  51.46 &
  52.91 &
  51.90 &
  52.54 &
  52.81 &
  52.06 &
  51.99 &
  52.24 &
  51.85 &
  52.98 &
  52.15 &
  52.18 &
  53.10 &
  52.38 &
  50.77 &
  52.20 \\
 &
  pt &
  56.38 &
  52.16 &
  55.22 &
  53.61 &
  52.58 &
  55.29 &
  53.94 &
  52.73 &
  53.65 &
  53.03 &
  55.44 &
  53.30 &
  51.47 &
  55.86 &
  54.36 &
  49.72 &
  53.31 \\
 &
  ru &
  55.94 &
  51.55 &
  54.60 &
  52.97 &
  52.92 &
  54.05 &
  53.87 &
  52.55 &
  53.22 &
  52.42 &
  54.62 &
  53.68 &
  51.76 &
  54.76 &
  54.76 &
  50.60 &
  53.23 \\
\multirow{-7}{*}{UNITER} &
  zh &
  53.12 &
  50.72 &
  52.19 &
  51.90 &
  52.04 &
  52.30 &
  51.60 &
  52.02 &
  51.82 &
  48.72 &
  49.71 &
  49.26 &
  48.36 &
  49.71 &
  48.94 &
  48.09 &
  48.97 \\ \hline
 &
  bn &
  54.16 &
  52.34 &
  53.32 &
  52.42 &
  52.16 &
  53.12 &
  52.96 &
  51.59 &
  52.56 &
  52.42 &
  53.06 &
  52.39 &
  50.84 &
  52.80 &
  52.57 &
  50.16 &
  52.03 \\
 &
  de &
  55.22 &
  52.23 &
  54.85 &
  53.43 &
  52.75 &
  54.26 &
  53.69 &
  52.22 &
  53.35 &
  52.58 &
  54.80 &
  52.84 &
  51.40 &
  54.13 &
  53.36 &
  49.73 &
  52.69 \\
 &
  id &
  55.33 &
  52.56 &
  54.52 &
  54.38 &
  53.46 &
  54.22 &
  54.08 &
  53.12 &
  53.76 &
  52.85 &
  54.13 &
  54.25 &
  51.87 &
  54.31 &
  53.57 &
  50.99 &
  53.14 \\
 &
  ko &
  53.62 &
  51.34 &
  52.93 &
  52.27 &
  52.35 &
  52.78 &
  52.45 &
  51.76 &
  52.27 &
  51.81 &
  52.76 &
  51.86 &
  51.81 &
  52.70 &
  52.12 &
  50.21 &
  51.90 \\
 &
  pt &
  55.88 &
  52.33 &
  54.77 &
  53.41 &
  52.35 &
  54.87 &
  53.93 &
  52.57 &
  53.46 &
  52.21 &
  54.33 &
  53.45 &
  51.36 &
  55.06 &
  53.42 &
  49.41 &
  52.75 \\
 &
  ru &
  55.12 &
  52.19 &
  53.94 &
  52.76 &
  52.59 &
  53.84 &
  54.17 &
  52.67 &
  53.17 &
  52.44 &
  54.18 &
  53.14 &
  51.04 &
  54.11 &
  54.01 &
  50.10 &
  52.72 \\
\multirow{-7}{*}{VILBERT} &
  zh &
  52.87 &
  50.53 &
  52.35 &
  51.71 &
  51.54 &
  52.10 &
  51.61 &
  51.49 &
  51.62 &
  48.56 &
  49.26 &
  48.78 &
  47.77 &
  49.03 &
  48.64 &
  48.13 &
  48.60 \\ \hline
 &
  bn &
  52.20 &
  50.13 &
  51.50 &
  50.58 &
  49.71 &
  50.98 &
  50.25 &
  50.02 &
  50.45 &
  50.84 &
  51.80 &
  50.65 &
  49.28 &
  51.47 &
  50.86 &
  48.46 &
  50.48 \\
 &
  de &
  53.51 &
  50.57 &
  53.10 &
  51.17 &
  50.45 &
  52.59 &
  51.47 &
  50.97 &
  51.47 &
  50.92 &
  53.29 &
  50.90 &
  49.10 &
  52.42 &
  51.60 &
  48.25 &
  50.93 \\
 &
  id &
  52.82 &
  50.38 &
  51.71 &
  52.01 &
  50.95 &
  52.16 &
  51.69 &
  51.04 &
  51.42 &
  51.03 &
  51.55 &
  51.67 &
  49.86 &
  51.97 &
  51.05 &
  49.07 &
  50.89 \\
 &
  ko &
  52.53 &
  50.06 &
  51.50 &
  50.68 &
  51.07 &
  51.70 &
  51.04 &
  50.75 &
  50.97 &
  50.44 &
  51.91 &
  50.63 &
  50.82 &
  51.83 &
  50.91 &
  49.59 &
  50.88 \\
 &
  pt &
  54.41 &
  50.72 &
  53.50 &
  51.86 &
  50.79 &
  53.72 &
  52.19 &
  51.45 &
  52.03 &
  51.17 &
  53.22 &
  51.32 &
  49.51 &
  53.74 &
  52.19 &
  48.66 &
  51.40 \\
 &
  ru &
  53.51 &
  50.14 &
  52.73 &
  51.34 &
  50.94 &
  52.35 &
  51.96 &
  51.04 &
  51.50 &
  51.30 &
  52.97 &
  51.73 &
  49.83 &
  52.93 &
  52.80 &
  49.47 &
  51.58 \\
\multirow{-7}{*}{VisualBERT} &
  zh &
  49.04 &
  47.69 &
  48.54 &
  48.02 &
  48.03 &
  48.48 &
  47.82 &
  47.96 &
  48.08 &
  47.44 &
  48.36 &
  48.16 &
  47.64 &
  48.64 &
  47.66 &
  46.96 &
  47.84 \\ \hline
 &
  bn &
  54.79 &
  52.91 &
  54.39 &
  52.70 &
  53.37 &
  53.72 &
  53.36 &
  52.81 &
  53.32 &
  53.30 &
  54.29 &
  52.91 &
  51.76 &
  53.80 &
  53.70 &
  51.35 &
  53.02 \\
 &
  de &
  55.61 &
  52.38 &
  55.27 &
  53.43 &
  52.58 &
  54.63 &
  53.32 &
  52.31 &
  53.42 &
  52.84 &
  55.06 &
  53.13 &
  51.47 &
  54.55 &
  53.78 &
  49.58 &
  52.92 \\
 &
  id &
  55.43 &
  52.40 &
  54.52 &
  54.16 &
  53.74 &
  54.14 &
  53.80 &
  53.08 &
  53.69 &
  53.04 &
  54.37 &
  54.09 &
  52.00 &
  54.30 &
  53.92 &
  50.95 &
  53.24 \\
 &
  ko &
  54.04 &
  51.79 &
  53.59 &
  52.75 &
  52.93 &
  53.40 &
  52.73 &
  52.02 &
  52.74 &
  52.00 &
  53.29 &
  52.62 &
  52.06 &
  53.17 &
  52.77 &
  50.45 &
  52.34 \\
 &
  pt &
  56.54 &
  52.56 &
  55.52 &
  53.77 &
  52.87 &
  55.62 &
  54.29 &
  52.41 &
  53.86 &
  53.10 &
  55.49 &
  53.55 &
  51.60 &
  55.81 &
  54.33 &
  49.91 &
  53.40 \\
 &
  ru &
  55.84 &
  52.11 &
  54.83 &
  53.67 &
  52.73 &
  54.65 &
  54.08 &
  52.71 &
  53.54 &
  53.27 &
  55.01 &
  53.71 &
  51.97 &
  54.47 &
  54.67 &
  50.87 &
  53.42 \\ 
\multirow{-7}{*}{VL-BERT} &
  zh &
  50.85 &
  48.76 &
  50.12 &
  49.52 &
  49.25 &
  49.99 &
  49.45 &
  49.37 &
  49.49 &
  48.82 &
  49.73 &
  49.32 &
  48.49 &
  49.83 &
  49.23 &
  48.62 &
  49.15 \\ \hline
 &
  bn &
  55.06 &
  52.97 &
  53.86 &
  53.57 &
  52.89 &
  53.60 &
  53.32 &
  52.60 &
  53.26 &
  53.78 &
  54.26 &
  53.63 &
  52.28 &
  53.99 &
  53.67 &
  51.06 &
  53.24 \\
 &
  de &
  56.11 &
  53.18 &
  55.70 &
  53.98 &
  53.51 &
  55.11 &
  54.25 &
  53.31 &
  54.15 &
  53.74 &
  56.07 &
  53.70 &
  51.96 &
  55.15 &
  54.37 &
  50.80 &
  53.68 \\
 &
  id &
  55.34 &
  52.54 &
  54.17 &
  54.29 &
  53.78 &
  53.76 &
  53.41 &
  52.94 &
  53.56 &
  53.44 &
  54.36 &
  54.46 &
  52.36 &
  54.06 &
  54.13 &
  50.82 &
  53.38 \\
 &
  ko &
  55.14 &
  52.51 &
  54.25 &
  53.18 &
  53.63 &
  53.73 &
  53.33 &
  53.09 &
  53.39 &
  53.17 &
  53.83 &
  53.37 &
  53.04 &
  54.21 &
  53.31 &
  52.12 &
  53.29 \\
 &
  pt &
  56.07 &
  52.78 &
  55.22 &
  53.93 &
  52.86 &
  55.29 &
  54.53 &
  53.39 &
  54.00 &
  52.97 &
  55.07 &
  53.42 &
  51.61 &
  55.72 &
  54.60 &
  50.44 &
  53.40 \\
 &
  ru &
  56.01 &
  52.66 &
  54.78 &
  53.58 &
  53.59 &
  54.52 &
  54.24 &
  53.29 &
  53.81 &
  54.05 &
  55.45 &
  53.83 &
  52.18 &
  54.94 &
  55.22 &
  51.68 &
  53.91 \\
\multirow{-7}{*}{BLIP-2} &
  zh &
  50.48 &
  48.63 &
  49.94 &
  49.74 &
  49.63 &
  49.55 &
  49.68 &
  49.89 &
  49.58 &
  49.25 &
  50.05 &
  49.40 &
  48.40 &
  49.90 &
  49.94 &
  49.22 &
  49.45 \\ \hline
 &
  bn &
  55.21 &
  52.92 &
  54.01 &
  53.38 &
  53.17 &
  53.93 &
  53.23 &
  53.00 &
  53.38 &
  53.68 &
  54.77 &
  53.75 &
  51.69 &
  54.38 &
  53.80 &
  51.65 &
  53.39 \\
 &
  de &
  55.84 &
  53.04 &
  55.06 &
  53.82 &
  53.17 &
  54.32 &
  54.08 &
  53.18 &
  53.81 &
  53.50 &
  55.42 &
  53.51 &
  51.93 &
  54.48 &
  53.78 &
  50.74 &
  53.34 \\
 &
  id &
  55.97 &
  52.96 &
  54.61 &
  54.46 &
  54.13 &
  54.64 &
  54.56 &
  53.67 &
  54.15 &
  53.67 &
  54.85 &
  54.86 &
  52.85 &
  54.73 &
  54.73 &
  51.67 &
  53.91 \\
 &
  ko &
  54.62 &
  51.93 &
  53.67 &
  52.91 &
  53.24 &
  53.27 &
  53.29 &
  53.16 &
  53.07 &
  52.76 &
  53.82 &
  53.08 &
  52.90 &
  53.39 &
  52.87 &
  51.63 &
  52.92 \\
 &
  pt &
  56.64 &
  52.85 &
  55.54 &
  54.07 &
  52.98 &
  55.61 &
  54.75 &
  53.45 &
  54.18 &
  53.77 &
  55.39 &
  53.78 &
  51.90 &
  56.09 &
  54.87 &
  50.67 &
  53.78 \\
 &
  ru &
  56.27 &
  52.39 &
  54.85 &
  53.50 &
  53.39 &
  54.34 &
  54.00 &
  52.94 &
  53.63 &
  53.81 &
  55.29 &
  54.13 &
  52.08 &
  54.87 &
  54.77 &
  51.15 &
  53.73 \\
\multirow{-7}{*}{InstructBLIP} &
  zh &
  51.66 &
  50.22 &
  51.06 &
  50.57 &
  50.46 &
  50.54 &
  50.33 &
  50.40 &
  50.51 &
  50.02 &
  51.03 &
  50.51 &
  49.43 &
  50.83 &
  50.21 &
  49.40 &
  50.20 \\ \hline
 &
  bn &
  55.70 &
  54.08 &
  55.14 &
  54.20 &
  54.09 &
  54.63 &
  54.60 &
  54.00 &
  54.39 &
  54.13 &
  54.70 &
  53.77 &
  52.67 &
  54.25 &
  54.06 &
  51.94 &
  53.65 \\
 &
  de &
  56.94 &
  53.97 &
  56.41 &
  54.99 &
  54.52 &
  55.95 &
  55.70 &
  53.87 &
  55.06 &
  54.08 &
  56.28 &
  54.01 &
  52.77 &
  55.66 &
  55.14 &
  51.30 &
  54.18 \\
 &
  id &
  56.26 &
  53.82 &
  55.36 &
  55.27 &
  54.63 &
  55.29 &
  54.86 &
  54.01 &
  54.75 &
  54.05 &
  55.40 &
  55.14 &
  52.97 &
  55.14 &
  54.54 &
  51.92 &
  54.17 \\
 &
  ko &
  55.42 &
  53.54 &
  54.42 &
  54.17 &
  54.43 &
  54.64 &
  54.25 &
  53.70 &
  54.16 &
  53.39 &
  54.40 &
  53.67 &
  53.65 &
  54.66 &
  53.91 &
  52.03 &
  53.67 \\
 &
  pt &
  57.17 &
  53.79 &
  56.07 &
  54.73 &
  53.67 &
  56.38 &
  55.10 &
  53.75 &
  54.78 &
  54.07 &
  55.91 &
  54.16 &
  52.50 &
  56.33 &
  54.76 &
  50.90 &
  54.09 \\
 &
  ru &
  55.57 &
  53.26 &
  54.95 &
  53.90 &
  53.61 &
  54.87 &
  54.79 &
  53.50 &
  54.13 &
  53.28 &
  54.79 &
  53.88 &
  52.19 &
  54.68 &
  54.88 &
  50.91 &
  53.52 \\
\multirow{-7}{*}{FLAVA} &
  zh &
  51.93 &
  49.79 &
  51.40 &
  50.85 &
  50.22 &
  51.11 &
  50.40 &
  50.76 &
  50.65 &
  49.89 &
  50.78 &
  50.12 &
  49.19 &
  50.64 &
  50.02 &
  49.59 &
  50.03 \\ \hline
 \toprule
\end{tabular}
\end{adjustbox}

\caption{Translate-test results with varied pivot languages during RT translation. The averaged results are in Table~\ref{tab:varied_pivot}.}
\label{tab:pivot_full}
\end{table*}

\noindent \textbf{Full results of Fig.~\ref{fig:maxm}}
Table~\ref{tab:maxm_full} presents the full results of different models on the MaXM dataset. \textsc{NLLB-200-3.3B} is used as an MT system for translate-test evaluation.

\begin{table*}[t!]
\small
\centering
\begin{tabular}{|cc|c|ccccccc|}
\hline
 & \multicolumn{1}{l}{} & \multicolumn{1}{l}{} & \multicolumn{7}{c|}{\textit{Translate-Test}} \\ \cline{4-10}
\textbf{Models} & \multicolumn{1}{c}{\textbf{RT?}} & \multicolumn{1}{c}{en} & fr & hi & ro & th & yi & zh & avg. \\ \hline
\multirow{2}{*}{MUNITER} &  & 61.73 & 60    & 72.73 & 63.33 & 65.79 & 55.13 & 53.85 & 61.81 \\
 & \checkmark& 58.02 & 64    & 76.14 & 62.22 & 61.84 & 46.15 & 65.38 & 62.62 \\ \hline
\multirow{2}{*}{XUNITER} &  & 60.49 & 61.33 & 79.55 & 65.56 & 57.89 & 44.87 & 63.46 & 62.11 \\
 & \checkmark& 54.32 & 62.67 & 79.55 & 62.22 & 56.58 & 48.72 & 61.54 & 61.88 \\ \hline
\multirow{2}{*}{UC$^2$} &  & 60.49 & 60.00 & 68.18 & 58.89 & 56.58 & 51.28 & 59.62 & 59.09 \\
 & \checkmark& 51.85 & 64.00 & 75.00 & 60.00 & 56.58 & 57.69 & 57.69 & 61.83 \\ \hline
 \multirow{2}{*}{M$^3$P} &  & 59.26 & 62.67 & 76.14 & 66.67 & 53.95 & 43.59 & 61.54 & 60.76 \\
 & \checkmark& 64.20 & 66.67 & 75.00 & 70.00 & 61.84 & 53.85 & 59.62 & 64.50 \\
 \hline
\multirow{2}{*}{LXMERT} &  & 64.2 & 72.00 & 75.00 & 61.11 & 52.63 & 46.15 & 55.77 & 60.44 \\
 & \checkmark& 64.20 & 72.00 & 79.55 & 65.56 & 61.84 & 55.13 & 63.46 & 66.26 \\ \hline
\multirow{2}{*}{UNITER} &  & 61.73 & 70.67 & 76.14 & 67.78 & 59.21 & 46.15 & 63.46 & 63.90 \\
 & \checkmark& 61.73 & 68    & 76.14 & 63.33 & 56.58 & 58.97 & 57.69 & 63.45 \\ \hline
\multirow{2}{*}{VILBERT} &  & 60.49 & 66.67 & 75.00 & 66.67 & 63.16 & 48.72 & 61.54 & 63.63 \\
 & \checkmark& 62.96 & 66.67 & 76.14 & 63.33 & 60.53 & 53.85 & 59.62 & 63.36 \\ \hline
\multirow{2}{*}{VisualBERT} &  & 69.14 & 70.67 & 71.59 & 60.00 & 56.58 & 51.28 & 59.62 & 61.62 \\
 & \checkmark& 70.37 & 70.67 & 76.14 & 60.00 & 68.42 & 57.69 & 65.38 & 66.38 \\ \hline
\multirow{2}{*}{VL-BERT} &  & 35.80 & 44.00 & 56.82 & 37.78 & 48.68 & 47.44 & 48.08 & 47.13 \\
 & \checkmark& 50.62 & 48.00 & 52.27 & 40.00 & 44.74 & 46.15 & 42.31 & 45.58 \\ \hline
 \multirow{2}{*}{BLIP-2} &  & 60.49 & 72.00 & 72.73 & 76.67 & 65.79 & 71.79 & 57.69 & 69.45 \\
 & \checkmark               & 60.49 & 72.00 & 69.32 & 70.00 & 64.47 & 67.95 & 67.31 & 68.51 \\ \hline
\multirow{2}{*}{InstructBLIP} &  & 64.20 & 73.33 & 73.86 & 72.22 & 65.79 & 71.79 & 57.69 & 69.11 \\
 & \checkmark                    & 62.96 & 69.33 & 69.32 & 70.00 & 61.84 & 65.38 & 61.54 & 66.24 \\ \hline
\multirow{2}{*}{FLAVA}  &  & 64.20 & 70.67 & 80.68 & 76.67 & 75.00 & 65.38 & 67.31 & 72.62 \\
 & \checkmark              & 65.43 & 76.00 & 76.14 & 75.56 & 75.00 & 67.95 & 71.15 & 73.63 \\ \hline
 
\end{tabular}

\caption{Full results on MaXM dataset. The averaged results across different models are in Table~\ref{fig:maxm}.}
\label{tab:maxm_full}
\end{table*}

\noindent \textbf{Full results of Table~\ref{tab:da_main}}
Table~\ref{tab:da_full} presents the full results of models with different data sources. 

\begin{table*}[t!]
\small
\centering
\begin{tabular}{|clc|c|ccccccc|}
\hline
 & \multicolumn{1}{l}{} & \multicolumn{1}{l}{} & \multicolumn{8}{c|}{\textit{Translate-Test}} \\ \cline{4-11}
\textbf{Models} & \multicolumn{1}{l}{} & \multicolumn{1}{c}{en} & bn & de & id & ko & pt & ru & zh & \textbf{avg.} \\ \hline
\multirow{4}{*}{MUNITER} & Human & 57.33 & 50.67 & 54.09 & 52.54 & 50.67 & 54.21 & 49.69 & 49.57 & 51.63 \\ 
 & MT & 55.70 & 52.34 & 55.66 & 53.48 & 53.36 & 54.72 & 53.98 & 52.29 & 53.69 \\
 & MERGE & 57.12 & 52.79 & 55.99 & 53.78 & 53.7 & 55.15 & 53.94 & 52.97 & 54.05 \\
 & TAG & 57.08 & 52.92 & 56.21 & 54.68 & 53.48 & 55.72 & 54.85 & 53.18 & 54.43 \\ \hline
\multirow{4}{*}{XUNITER} & Human & 56.98 & 50.76 & 54.63 & 52.37 & 50.52 & 54.24 & 48.91 & 49.94 & 51.62 \\ 
 & MT & 55.22 & 52.10 & 54.97 & 52.66 & 52.51 & 54.18 & 52.85 & 52.23 & 53.07 \\
 & MERGE & 56.69 & 52.5 & 55.45 & 53.55 & 53.07 & 54.83 & 53.71 & 52.61 & 53.67 \\
 & TAG & 56 & 52.2 & 55.00 & 53.12 & 52.62 & 54.61 & 53.54 & 52.04 & 53.30 \\ \hline
\multirow{4}{*}{UC$^2$} & Human & 56.85 & 51.34 & 54.01 & 52.35 & 50.75 & 53.81 & 51.93 & 50.04 & 52.03 \\ 
 & MT & 55.12 & 52.35 & 55.10 & 53.29 & 53.07 & 54.17 & 53.36 & 52.73 & 53.44 \\
 & MERGE & 57.67 & 53.84 & 56.59 & 54.87 & 54.48 & 56.11 & 55.08 & 53.45 & 54.92 \\
 & TAG & 56.7 & 53.24 & 55.95 & 54.01 & 53.59 & 55.48 & 54.95 & 53.11 & 54.33 \\ \hline
 \multirow{4}{*}{M$^3$P} & Human & 54.45 & 49.18 & 52.14 & 49.87 & 48.59 & 51.87 & 49.05 & 48.38 & 49.87 \\
 & MT & 52.97 & 50.63 & 53.03 & 51.42 & 50.38 & 52.11 & 51.80 & 50.41 & 51.40 \\
 & MERGE & 53.7 & 50.37 & 52.85 & 50.89 & 50.36 & 51.88 & 51.22 & 50.43 & 51.14 \\
 & TAG & 54.66 & 51.11 & 53.71 & 51.66 & 50.78 & 53.12 & 52.38 & 51.27 & 52.00 \\
 \hline
\multirow{4}{*}{LXMERT} & Human & 55.40 & 49.64 & 52.83 & 50.80 & 49.17 & 52.49 & 47.54 & 48.02 & 50.07 \\ 
 & MT & 53.44 & 50.20 & 52.93 & 51.34 & 50.41 & 52.47 & 51.44 & 50.25 & 51.29 \\
 & MERGE & 54.88 & 50.78 & 53.59 & 52.01 & 51.28 & 53.04 & 52.31 & 50.68 & 51.96 \\
 & TAG & 54.75 & 51.03 & 53.82 & 52.17 & 51.26 & 53.24 & 52.15 & 51.2 & 52.12 \\ \hline
\multirow{4}{*}{UNITER} & Human & 57.47 & 51.74 & 54.52 & 52.79 & 51.27 & 54.56 & 52.27 & 50.33 & 52.50 \\
 & MT & 55.92 & 52.32 & 55.53 & 53.67 & 52.93 & 54.66 & 53.56 & 52.6 & 53.61 \\
 & MERGE & 57.26 & 52.97 & 56.19 & 54.05 & 53.65 & 55.53 & 54.44 & 53.1 & 54.28 \\
 & TAG & 57.03 & 52.71 & 55.96 & 54.21 & 53.16 & 55.45 & 54.48 & 52.74 & 54.10 \\ \hline
\multirow{4}{*}{VILBERT} & Human & 56.72 & 50.84 & 54.10 & 52.27 & 50.73 & 53.98 & 49.91 & 49.92 & 51.68 \\
 & MT & 55.22 & 52.23 & 54.85 & 53.43 & 52.75 & 54.26 & 53.69 & 52.22 & 53.35 \\
 & MERGE & 56.97 & 53.01 & 55.46 & 53.73 & 53.54 & 55.05 & 54.33 & 53.05 & 54.02 \\
 & TAG & 56.67 & 53.04 & 55.72 & 54.21 & 53.73 & 55.42 & 54.65 & 52.84 & 54.23 \\ \hline
\multirow{4}{*}{VisualBERT} & Human & 55.17 & 49.43 & 52.58 & 50.34 & 48.66 & 52.72 & 50.50 & 48.89 & 50.45 \\
 & MT & 53.51 & 50.57 & 53.10 & 51.17 & 50.45 & 52.59 & 51.47 & 50.97 & 51.47 \\
 & MERGE & 54.79 & 51.07 & 53.43 & 51.91 & 51.36 & 53.15 & 52.19 & 51.49 & 52.09 \\
 & TAG & 54.92 & 51.28 & 53.91 & 52.15 & 51.07 & 53.70 & 51.2 & 51.33 & 52.09 \\ \hline
\multirow{4}{*}{VL-BERT} & Human & 57.79 & 51.22 & 54.47 & 52.62 & 50.94 & 54.79 & 51.17 & 50.02 & 52.18 \\
 & MT & 55.61 & 52.38 & 55.27 & 53.43 & 52.58 & 54.63 & 53.32 & 52.31 & 53.42 \\
 & MERGE & 57.45 & 52.71 & 55.80 & 53.49 & 53.62 & 54.88 & 54.09 & 52.17 & 53.82 \\
 & TAG & 57.49 & 53.63 & 56.16 & 54.25 & 53.8 & 55.82 & 54.65 & 53.24 & 54.51 \\ \hline
 \multirow{4}{*}{BLIP-2} & Human & 58.05 & 52.03 & 54.70 & 52.99 & 51.57 & 54.91 & 52.36 & 51.22 & 52.83 \\
 & MT & 56.11 & 53.18 & 55.70 & 53.98 & 53.51 & 55.11 & 54.25 & 53.31 & 54.15 \\
 & MERGE & 57.41 & 53.6 & 56.26 & 54.33 & 53.83 & 55.94 & 54.52 & 53.76 & 54.61 \\
 & TAG & 57.31 & 53.62 & 56.23 & 54.33 & 53.98 & 55.72 & 55.14 & 53.78 & 54.69 \\ \hline
\multirow{4}{*}{InstructBLIP} & Human & 57.85 & 51.80 & 54.91 & 53.01 & 51.29 & 54.85 & 53.16 & 51.34 & 52.91 \\
 & MT & 55.84 & 53.04 & 55.06 & 53.82 & 53.17 & 54.32 & 54.08 & 53.18 & 53.81 \\
 & MERGE & 58.1 & 54.26 & 57.08 & 55.16 & 54.15 & 56.27 & 55.59 & 54.18 & 55.24 \\
 & TAG & 58.24 & 54.65 & 57.20 & 55.06 & 54.52 & 56.69 & 55.79 & 54.32 & 55.46 \\ \hline
\multirow{4}{*}{FLAVA} & Human & 58.84 & 53.47 & 56.26 & 54.11 & 52.85 & 55.84 & 53.64 & 52.18 & 54.05 \\
 & MT & 56.87 & 53.94 & 56.35 & 54.99 & 54.51 & 55.96 & 55.61 & 53.82 & 55.03 \\
 & MERGE & 57.95 & 53.95 & 56.61 & 54.99 & 54.33 & 56.08 & 55.22 & 53.91 & 55.01 \\
 & TAG & 57.44 & 54.21 & 56.56 & 55.18 & 54.51 & 55.95 & 55.42 & 53.73 & 55.08 \\ \hline
\end{tabular}
\caption{Full results of data augmentation experiments. The averaged results across different models are in Table~\ref{tab:da_main}.}
\label{tab:da_full}
\end{table*}

\begin{figure*}[ph]
\noindent\makebox[\textwidth]{%
\includegraphics[width=0.95\textwidth]{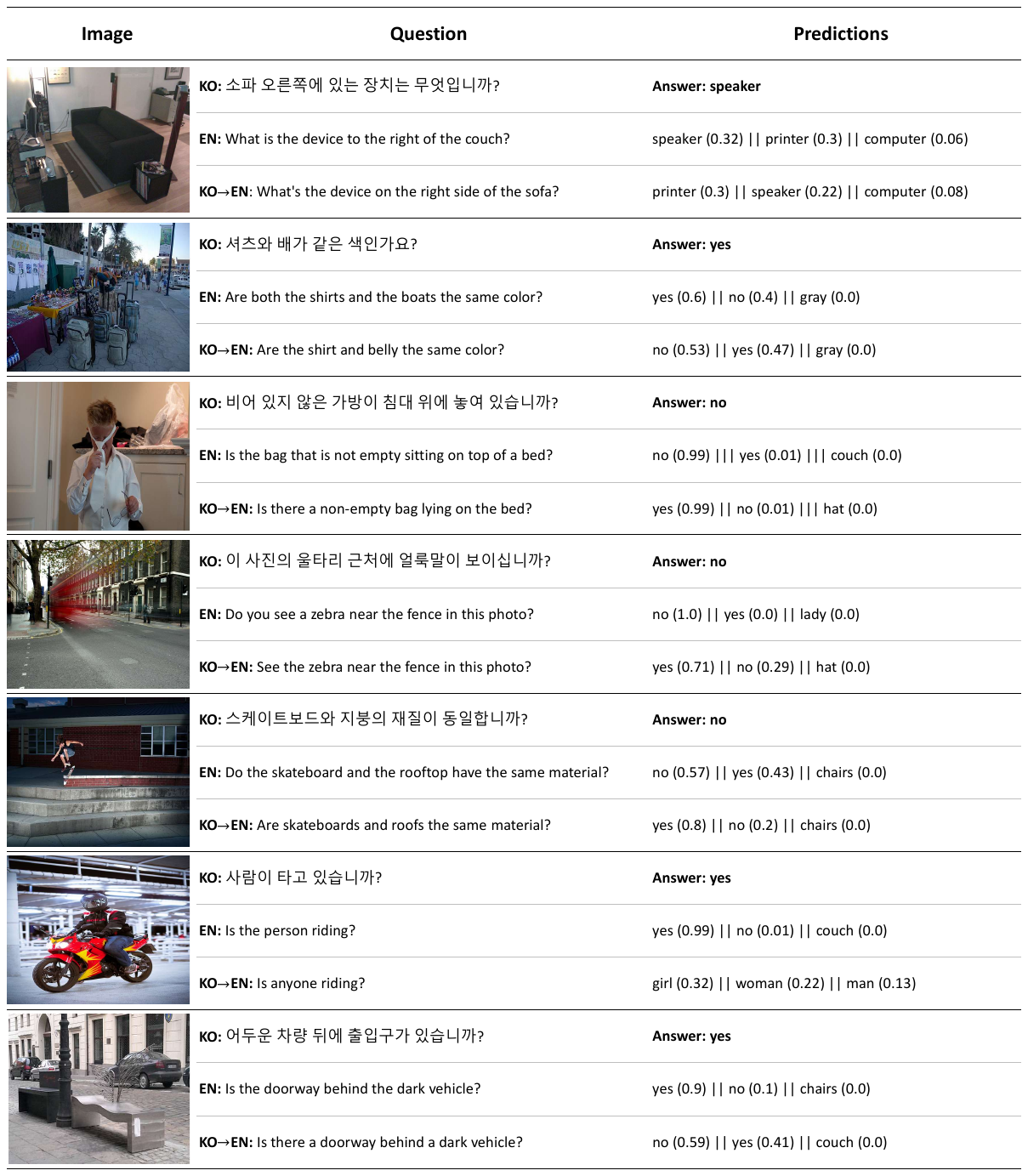}}
\vspace{-25mm}
\caption{
We present a randomly selected example, which includes the original English text (\textbf{EN}), its target language translation by a human annotator (e.g., \textbf{KO}), and translation from the target language to English (e.g., \textbf{KO $\rightarrow$ EN}) for translate-test. For each example, we provide the correct English answer, the top three English predictions, and the top three predictions from the translate-test, along with their respective probabilities of UC$^2$. In the translate-test, the examples with translation errors are specifically identified, with the type of error highlighted in \textcolor{red}{\textbf{red}}.
}
\label{fig:case_study_results}
\end{figure*}

\begin{figure*}[ph]
\noindent\makebox[\textwidth]{%
\includegraphics[width=0.95\textwidth]{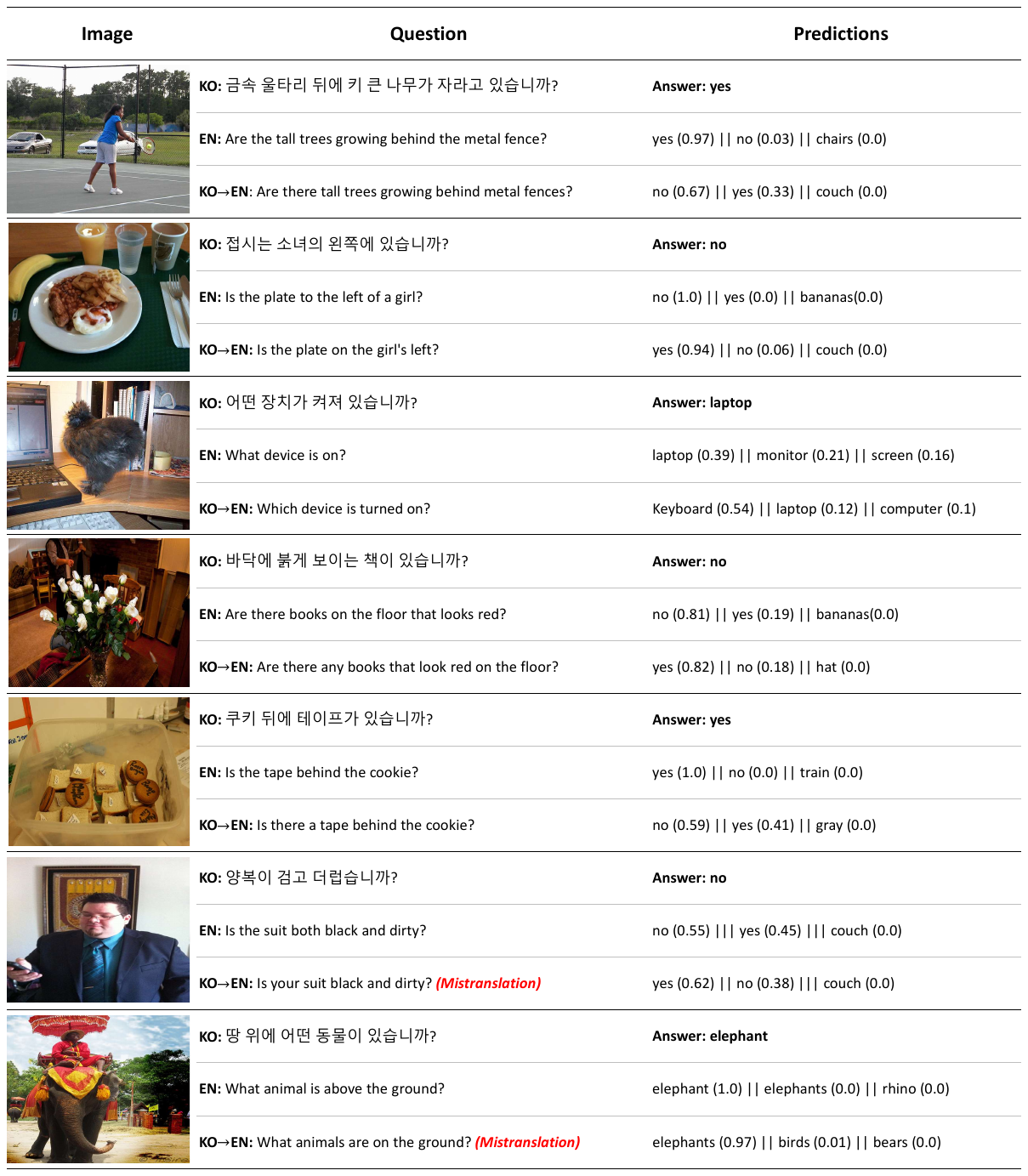}}
\vspace{-25mm}
\caption{
\textit{(cont'd)} We present a randomly selected example, which includes the original English text (\textbf{EN}), its target language translation by a human annotator (e.g., \textbf{KO}), and translation from the target language to English (e.g., \textbf{KO}) for translate-test. For each example, we provide the correct English answer, the top three English predictions, and the top three predictions from the translate-test, along with their respective probabilities of UC$^2$. In translate-test, examples with translation errors are specifically identified, with the type of error highlighted in \textcolor{red}{\textbf{red}}.
}
\label{fig:case_study_results_1}
\end{figure*}

\begin{figure*}[ph]
\noindent\makebox[\textwidth]{%
\includegraphics[width=0.95\textwidth]{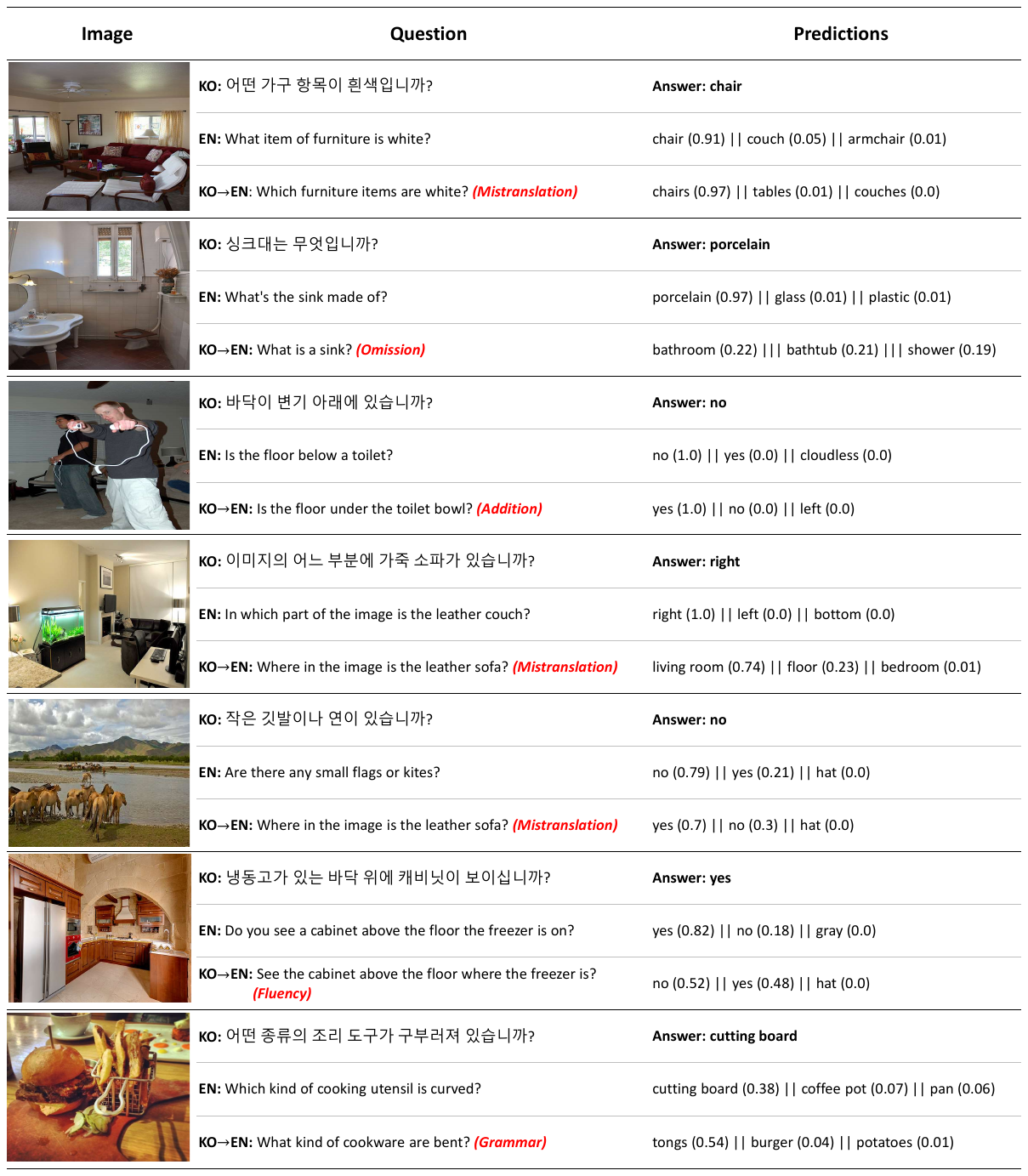}}
\vspace{-25mm}
\caption{ 
\textit{(cont'd)} We present a randomly selected example, which includes the original English text (\textbf{EN}), its target language translation by a human annotator (e.g., \textbf{KO}), and translation from the target language to English (e.g., \textbf{KO $\rightarrow$ EN}) for translate-test. For each example, we provide the correct English answer, the top three English predictions, and the top three predictions from the translate-test, along with their respective probabilities of UC$^2$. In translate-test, examples with translation errors are specifically identified, with the type of error highlighted in \textcolor{red}{\textbf{red}}.
}
\label{fig:case_study_results_2}
\end{figure*}

\begin{figure*}[ph]
\noindent\makebox[\textwidth]{%
\includegraphics[width=0.95\textwidth]{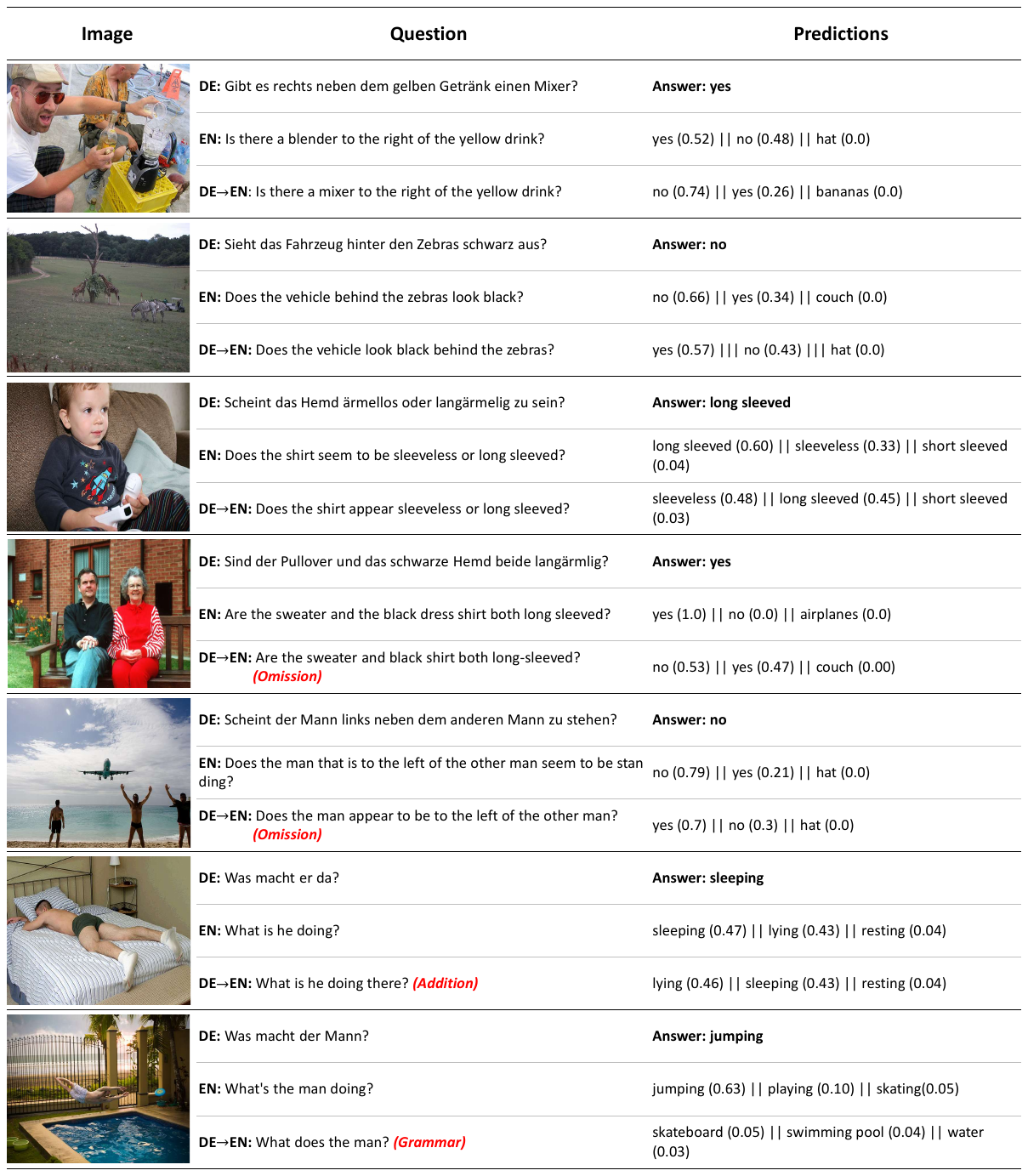}}
\vspace{-25mm}
\caption{ 
\textit{(cont'd)} We present a randomly selected example, which includes the original English text (\textbf{EN}), its target language translation by a human annotator (e.g., \textbf{DE}), and translation from the target language to English (e.g., \textbf{DE $\rightarrow$ EN}) for translate-test. For each example, we provide the correct English answer, the top three English predictions, and the top three predictions from the translate-test, along with their respective probabilities of UC$^2$. In translate-test, examples with translation errors are specifically identified, with the type of error highlighted in \textcolor{red}{\textbf{red}}.
}
\label{fig:case_study_results_3}
\end{figure*}

\begin{figure*}[pt]
\noindent\makebox[\textwidth]{%
\includegraphics[width=0.95\textwidth]{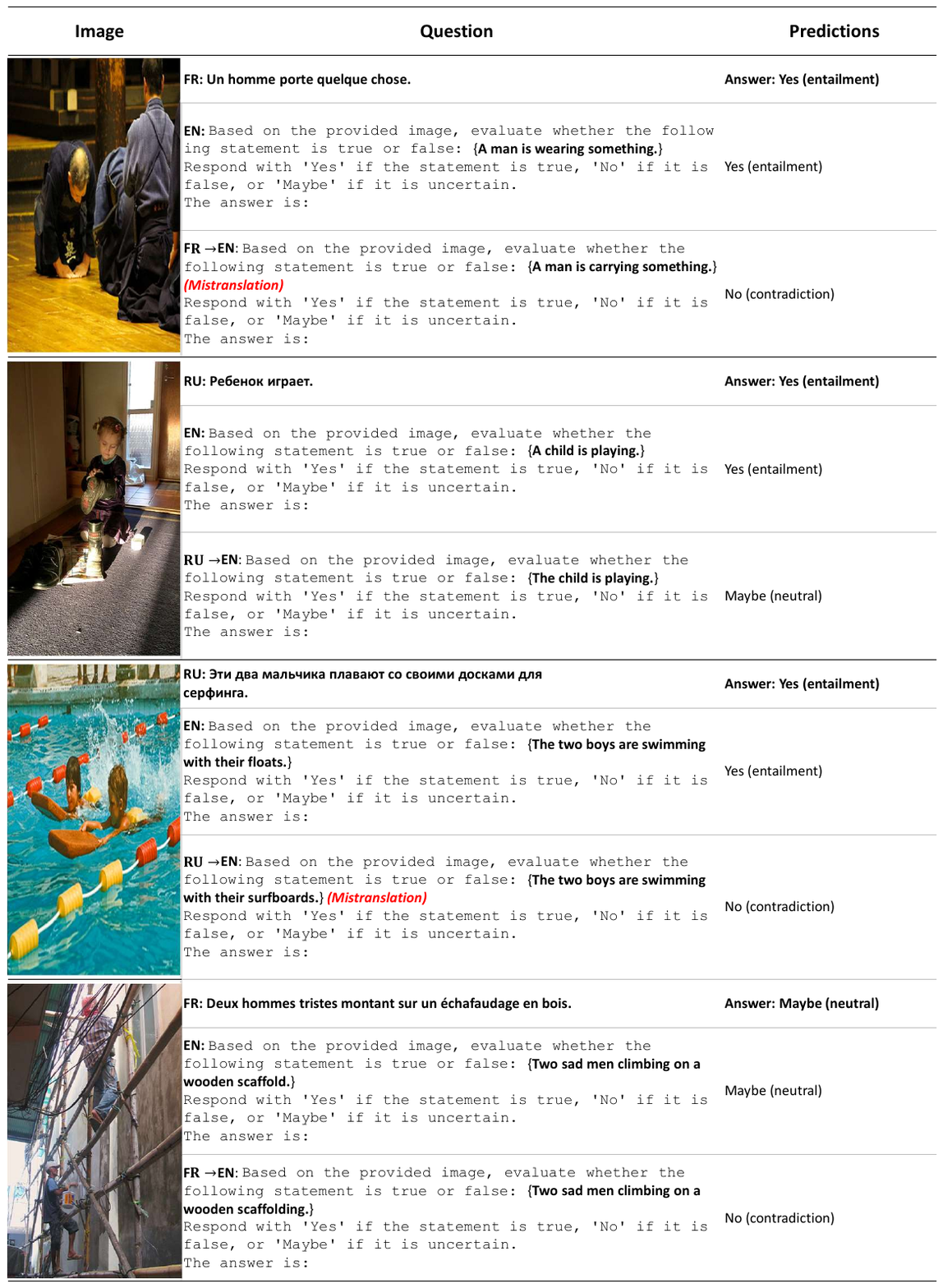}}
\vspace{-10mm}
\caption{
Sample results with \url{gpt-4-1106-vision-preview}. For each example, we present the original question written in the target language along with its answer (e.g., \textbf{FR}), the original question written in English and corresponding model prediction~(i.e., \textbf{EN}), and the translated question from the target language and model prediction~(e.g., \textbf{FR $\rightarrow$ EN}). Each question is given with a task description and is highlighted in \textbf{bold}. Any translation errors in translated questions are further highlighted in \textcolor{red}{red}.
}
\label{tab:gpt4_vision}
\end{figure*}


\begin{figure*}[pt]
\noindent\makebox[\textwidth]{%
\includegraphics[width=0.95\textwidth]{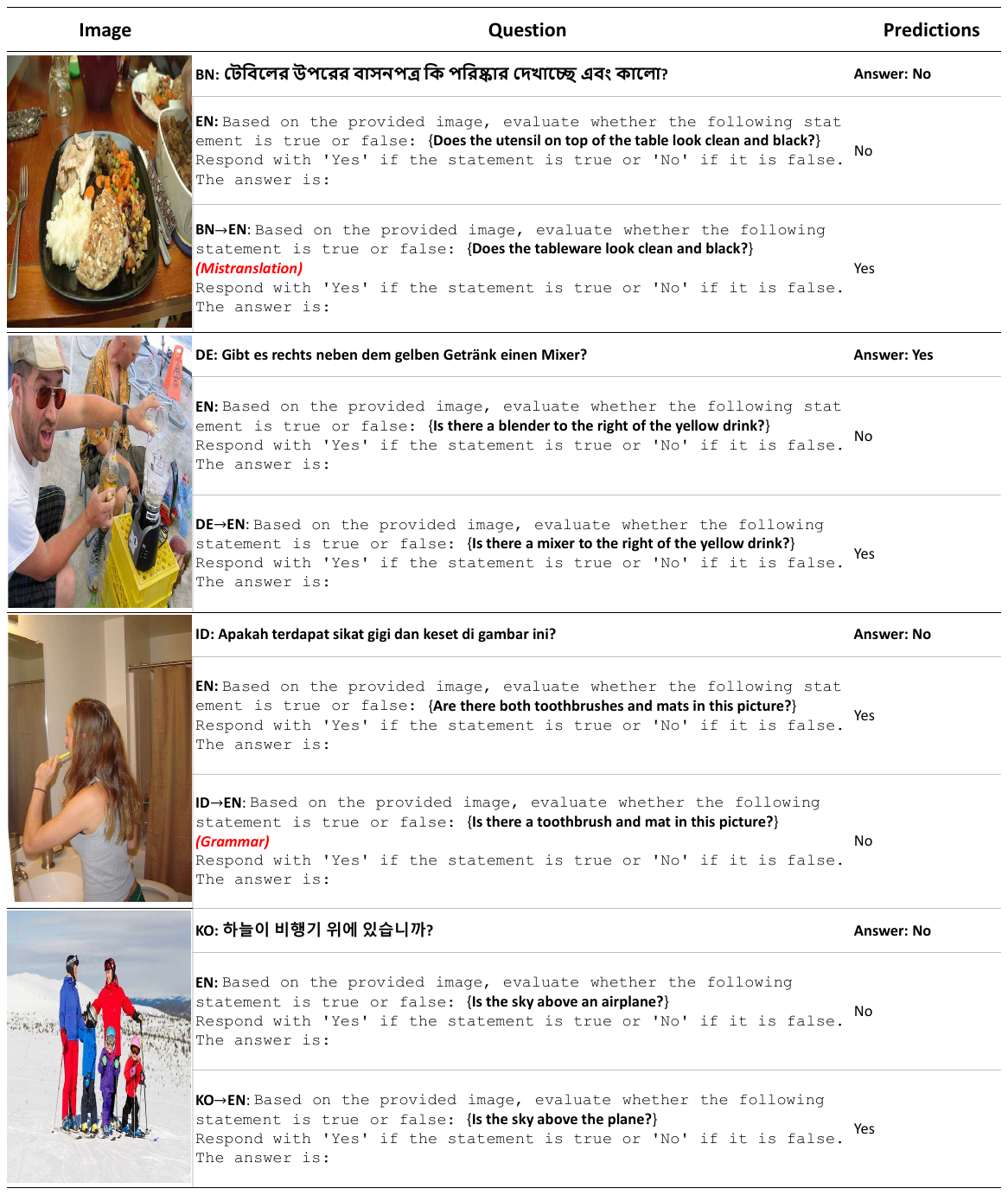}}
\vspace{-20mm}
\caption{
\textit{(cont'd)} Sample results with \url{gpt-4-1106-vision-preview}. For each example, we present the original question written in the target language along with its answer (e.g., \textbf{BN}), the original question written in English and corresponding model prediction~(i.e., \textbf{EN}), and the translated question from the target language and model prediction~(e.g., \textbf{BN $\rightarrow$ EN}). Each question is given with a task description and is highlighted in \textbf{bold}. Any translation errors in translated questions are further highlighted in \textcolor{red}{red}.
}
\label{tab:gpt4_vision_continued}
\end{figure*}

\end{document}